\documentclass[conference]{IEEEtran}
\usepackage{times}

\usepackage[numbers]{natbib}
\usepackage{multicol}
\usepackage{xcolor}
\definecolor{myblue}{rgb}{0.153, 0.66, 0.88}
\usepackage[bookmarks=true]{hyperref}
\hypersetup{
    colorlinks=true,
    linkcolor=myblue,
    filecolor=magenta,      
    urlcolor=myblue,
    citecolor=myblue,
}
\usepackage{graphicx}

\usepackage{algorithm}
\usepackage[noend]{algorithmic}
\usepackage{wrapfig}
\usepackage{amssymb}
\usepackage[normalem]{ulem}
\usepackage{mathtools}
\usepackage{booktabs}
\usepackage{subcaption}

\usepackage{lipsum}

\definecolor{mygreen}{rgb}{0.12, 0.54, 0.30}
\definecolor{myred}{rgb}{0.78, 0.33, 0.37}
\definecolor{myorange}{rgb}{0.98, 0.65, 0.15}
\newcommand{\algocap}{Behavior Retrieval}
\newcommand{\algo}{behavior retrieval}
\newcommand{\dt}{\mathcal{D}_t}
\newcommand{\dpr}{\mathcal{D}_{\text{prior}}}
\newcommand{\dr}{\mathcal{D}_{\text{ret}}}

\begin{document}

\title{
Behavior Retrieval: Few-Shot Imitation Learning \\ by Querying Unlabeled Datasets}

\author{Maximilian Du, Suraj Nair, Dorsa Sadigh, Chelsea Finn \\ Stanford University \\ \texttt{\{maxjdu, surajn\}@stanford.edu} }

\maketitle

\begin{abstract}
Enabling robots to learn novel visuomotor skills in a data-efficient manner remains an unsolved problem with myriad challenges. A popular paradigm for tackling this problem is through leveraging large unlabeled datasets that have many behaviors in them and then adapting a policy to a specific task using a small amount of task-specific human supervision (i.e. interventions or demonstrations). However, how best to leverage the narrow task-specific supervision and balance it with offline data remains an open question. 
Our key insight in this work is that task-specific data not only provides new data for an agent to train on but can also \emph{inform the type of prior data the agent should use for learning.}
Concretely, we propose a simple approach that uses a small amount of downstream expert data to selectively query relevant behaviors from an offline, unlabeled dataset (including many sub-optimal behaviors). The agent is then jointly trained on the expert and queried data.
We observe that our method learns to query only the relevant transitions to the task, filtering out sub-optimal or task-irrelevant data. By doing so, it is able to  learn more effectively from the mix of task-specific and offline data compared to na\"ively mixing the data or only using the task-specific data. 
Furthermore, we find that our simple querying approach outperforms more complex goal-conditioned methods by 20\% across simulated and real robotic manipulation tasks from images. See~\url{https://sites.google.com/view/behaviorretrieval} for videos and code.

\end{abstract}

\IEEEpeerreviewmaketitle

\section{Introduction}

\begin{figure}[t]
    \centering
    \includegraphics[width=0.99\linewidth]{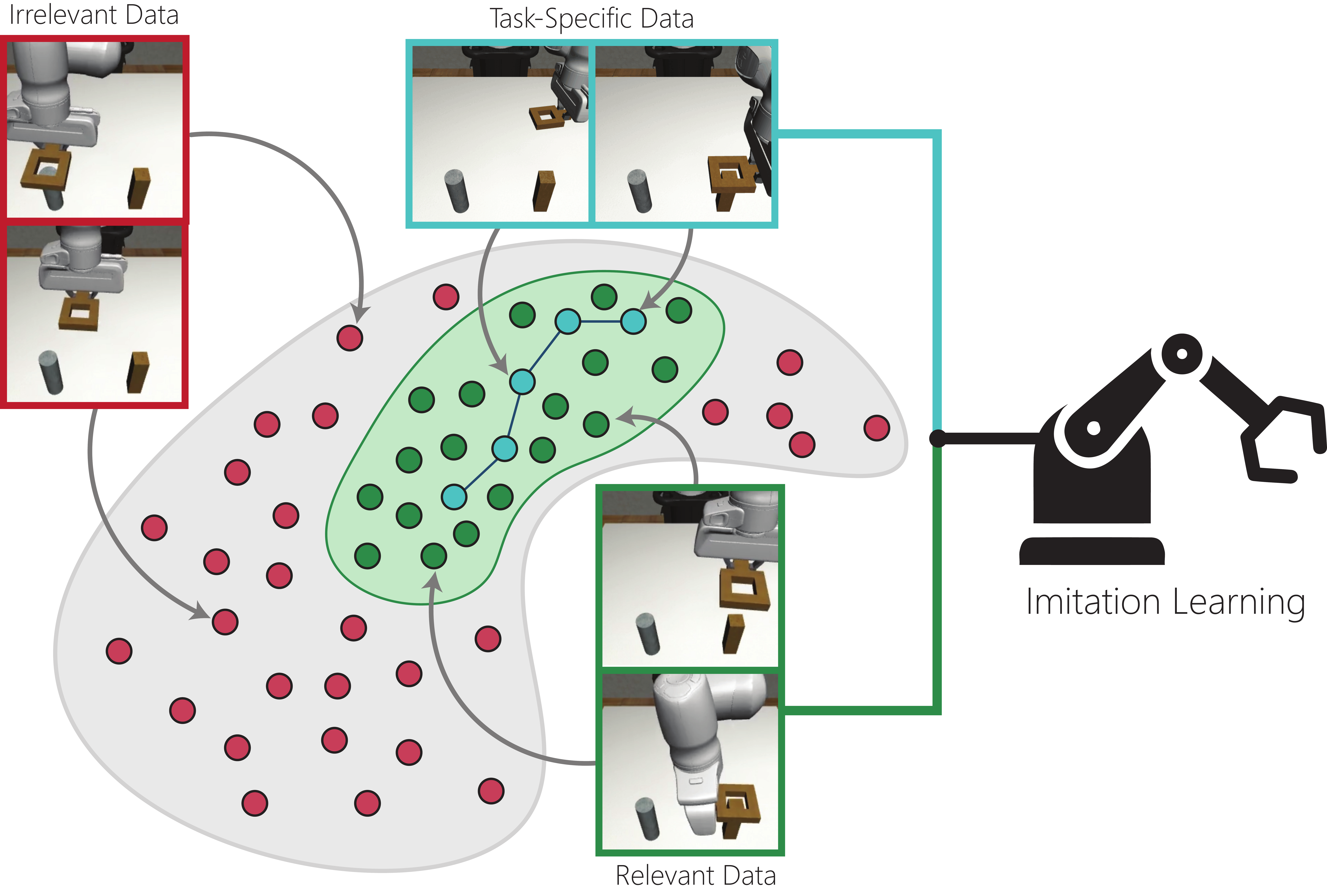}
    \vspace{-0.3cm}
    \caption{\small \textbf{Using Task-Specific Data to Query Offline Datasets}. Using a small amount of task-specific human expert feedback (i.e. interventions or demonstrations) (\textcolor{myblue}{blue}), our method learns to select the relevant portions (\textcolor{mygreen}{green}) of a broad, unlabeled offline dataset (\textcolor{myred}{red}) to efficiently learn the target task. In this example the task is to have the robot place the square on the right peg (shown with the initial and final frame in \textcolor{myblue}{blue}). While the broader dataset might include irrelevant data (initial and final frame shown in \textcolor{myred}{red}) where the robot is placing the square on the left peg, it includes useful data providing diversity in terms of how the square needs to be placed on the right peg (shown with initial and final frame in \textcolor{mygreen}{green}). Our algorithm identifies these relevant data from the broader dataset and learns from them while ignoring the irrelevant data. 
    }
    \vspace{-0.3cm}
    \label{pullfig}
\end{figure}

One of the promises of deep learning applied to robotics is the ability to learn control from sensor observations in a data-efficient manner, a prerequisite to deploying robots in rich real-world settings like homes.  
Towards accomplishing this goal, a number of prior methods have explored leveraging broad, unlabeled data to capture general knowledge and a small amount of task-specific data to adapt to a particular task. 
A natural form for this task-specific data to take is human expert data (e.g. demonstrations or online interventions), which can be imitated to efficiently adapt the agent to the target task. 
While this data is convenient to learn from, it comes at a cost: expensive human-in-the-loop supervision for every target task. Thus, even with pre-training data providing general knowledge, learning a performant task-specific policy without substantial human effort remains an open challenge.

Unlike the many works that study different techniques for offline pre-training from sub-optimal data, from visual pre-training \cite{Parisi2022TheUE, Nair2022R3MAU, Xiao2022MaskedVP} to multi-task imitation \cite{jang_bcz, Ebert2021BridgeDB, Shridhar2021CLIPortWA}, in this work we focus instead on how to 
best use the sub-optimal offline data in fine-tuning a policy from small amount of expert data.
Critically, most prior work do not make use of the offline data during fine-tuning, and those that do \cite{Ebert2021BridgeDB} tend to use simple heuristics that often fail to effectively balance offline and task-specific expert data, or focus on using the offline data for skills instead of low-level control \cite{Nasiriany2022LearningAR}. 
The interesting question to ask is \emph{why balancing these data sources tend to be difficult}. On one hand, fine-tuning the policy only on a small amount of downstream expert data can be unstable, diverging significantly from the pre-trained policy and losing much of the knowledge gained during pre-training. On the other hand, jointly training the agent on the online expert data and the offline pre-training data can be highly sub-optimal, as much of the pre-training data may be irrelevant or damaging to learning the target task. We study precisely this tradeoff, and seek to understand how best to balance task-specific human expert data and broad unlabeled data in learning control policies. 

Our key insight in this work is that the small amount of task-specific human expert data not only provides new data for an agent to train on but also \textbf{informs what prior data the agent should retrieve for learning}.
That is, when a human provides supervision to the agent through for example an intervention (a state-action tuple), the agent should not solely fine-tune on the tuple. Rather, the agent can learn to retrieve many relevant state-action tuples from a broad offline dataset and instead fine-tune on the larger combined data.

Concretely, we propose \algo, a method for training a policy on a small amount of human expert data (interventions or demonstrations), while simultaneously leveraging a broad, unlabeled dataset that may include
sub-optimal, task-irrelevant, and task-adversarial behaviors.
Our method begins by using the unlabeled dataset to pre-train a state-action similarity metric measuring the similarity between any two state-action pairs. Then, when adapting the policy with a small amount of human expert data, we retrieve relevant experience from the unlabeled dataset using our learned similarity metric, and jointly fine-tune the policy on the expert and queried data (See Figure~\ref{pullfig}). In a set of simulated and real-world robotic manipulation tasks from both low-dimensional states and image observations, 
we find that selectively sampling
from offline data for joint fine-tuning with our method leads to more stable and performant learning than only finetuning on the human expert data or using hand-designed heuristics for balancing offline and task-specific data.
Moreover, we find our proposed approach outperforms multi-task pre-training and finetuning 
  by over 20\% by using its simple retrieval strategy.

\section{Related Work}
\label{sec-rw}

Towards the goal of data-efficient policy learning, there are many relevant works spanning few-shot learning from experts which studies learning with just a few demos, offline pre-training which aims to learn priors from broad datasets, and meta-learning which learns policies that can adapt efficiently. 

\smallskip \textbf{Few-shot Learning from Experts.} Learning behaviors from human expert data through imitation learning is a long-studied problem in robotics \cite{Pomerleau-1989-15721, schall04, Ross2011ARO, Kelly2018HGDAggerII, parnichkun_reil, ARGALL2009469, imitationsurvey1, imitationsurvey2}. More recently, the emphasis for such imitation learning methods has been on learning efficiently, minimizing the effort placed on the human  expert. For example, many works have explored how a policy might be pre-trained on demonstrations then fine-tuned with online expert feedback \cite{Ross2011ARO, Laskey2017DARTNI, Mandlekar2020HumanintheLoopIL, Kelly2018HGDAggerII,  spencer2020, Jang2022BCZZT}. Similarly, many works have explored how to learn from broad data with many tasks and fine-tune on a new task given a small amount task-specific data \cite{Ebert2021BridgeDB,Mandi2021TowardsMG}. Rather than simply fine-tuning on expert interventions or demos, our method leverages \emph{retrieval}, filtering only the relevant prior experience and jointly training on the combined data. In our experiments, we compare to all of these prior methods and find that our approach \algo~outperforms them by selectively retrieving and training on offline data.

Additionally, several works have explored designing action spaces or planning trajectories in a way that enables more efficient learning \cite{zeng2020transporter, johns2021coarse-to-fine, qattentn, James2022CoarsetofineQW}. Behavior retrieval instead proposes a method for sampling from offline data, and in principle can be combined with any such action spaces in a straightforward manner. 
Recent approaches also attempt to be transparent about what type of data the imitation learning algorithm requires and further influence the demonstrator through interactive data collection to provide informative, in-distribution data \cite{Gandhi2022ElicitingCD, schrum2022reciprocal}. Such methods are complementary to our approach, and the two could be naturally combined to influence the demonstrator for better interventions which then enables even more effective retrieval from offline data using \algo.
Finally, our work is not the first to consider retrieval in imitation learning. 
Nasiriany et al. has explored learning skills from offline data and retrieving relevant skills for long-horizon tasks. Unlike this work, we learn low-level control policies, and focus on retrieving individual state-action tuples rather than high-level skills~\cite{Nasiriany2022LearningAR}. This allows us to leverage offline data to learn more performant low-level policies, as opposed to being tied to a high-level skill space. Other works like Goyal et al. \cite{Goyal2022RetrievalAugmentedRL} also explore the use of retrieval, however focus on sample efficient RL using retrieval from the replay buffer, rather than imitation from unstructured datasets like we consider in this work.   

\smallskip \textbf{Robotic Pre-training.} While our approach focuses on retrieving behaviors as a way of leveraging offline data, there exist many alternative methods for pre-training from offline data. For example, one line of work studies pre-training visual representations \cite{Parisi2022TheUE, Nair2022R3MAU, Xiao2022MaskedVP, Radosavovic2022, Ma2022VIPTU} that can then be used frozen or finetuned for more efficient downstream imitation or reinforcement learning. This approach is particularly effective at using data that does not contain action labels, like videos of humans. Another line of work focuses on using robot data with actions, and pre-training multi-task (e.g. goal-conditioned or language conditioned) policies through imitation learning \cite{dinggcbc, Lynch2019LearningLP, Belkhale2022PLATOPL, Brohan2022RT1RT, Ebert2021BridgeDB, jang_bcz}
 or offline reinforcement learning \cite{Kumar2022PreTrainingFR}. Like our work, some prior work also explores selectively sampling or weighting expert data \cite{Zhang2021ConfidenceAwareIL, beliaev2022imitation}. However, the focus of these works is on learning from a mix of sub-optimal experts for a particular task, while we instead focus on retrieving from a broad offline dataset during task-specific fine-tuning.
In total, all of these works are complementary to the direction we study in this work, as 
any of the above pre-training schemes can leverage the \algo~approach to efficiently finetune.

\smallskip \textbf{Meta-Imitation Learning.} Towards the goal of few-shot learning of control policies for new tasks and environments, another well-studied approach is that of meta-learning. Meta-imitation learning methods \cite{Finn2017OneShotVI, duan2017one, james2018task, yu2018one, zhou2019watch} train to produce effective policies using a few demonstrations of many tasks,
such that given just a few demos of a new task they can perform well. Critically all of these methods require a large dataset of well-defined, labeled tasks for meta-training, which can be expensive. Our approach is instead able to learn from unlabeled ``play'' data without annotated task boundaries and use it for few-shot imitation learning.

\section{\algocap: Imitation Learning by Querying Unlabeled Datasets} 
\label{sec-method}

\subsection{Problem Setup} 

\begin{figure*}[t]
    \centering
    \includegraphics[width=0.99\linewidth]{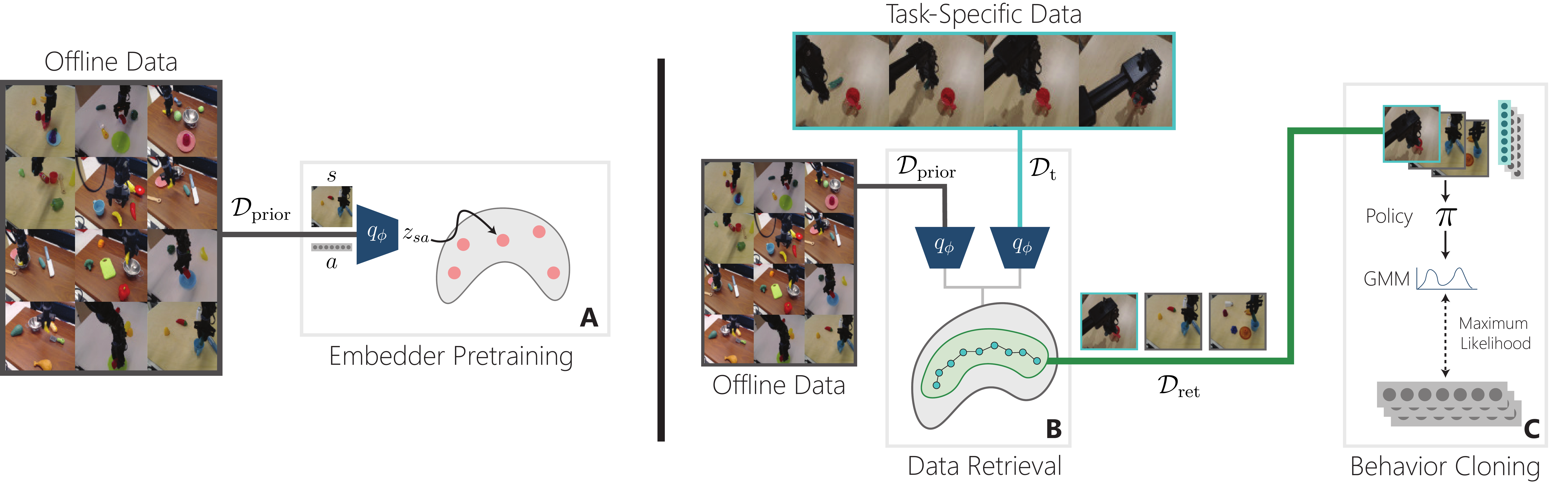}
    \vspace{-0.2cm}
    \caption{\small \textbf{The \algocap~Method}. Our approach has 3 main steps. \textbf{(A)} Using the unlabeled offline data $\dpr$ we pre-train a state-action embedding. \textbf{(B)} We use the pre-trained embedding to look up similar transitions in the offline data $\dpr$ that are relevant to the task data $\dt$. \textbf{(C)} We then train a policy with behavior cloning on the mix of the task-specific and retreived data.
    }
    \vspace{-0.6cm}
    \label{methodoverview}
\end{figure*}

In this work, we consider the problem setting of learning a target task from a small amount of task-specific expert data and a large amount of sub-optimal, unlabeled data, where all the data shares a state space $\mathcal{S}$ and an action space $\mathcal{A}$. Concretely, for a particular target task $t$ in the space of all tasks $\mathcal{T}$, we assume access to:
\begin{enumerate}
    \item $\dt$: a dataset of expert $(s, a)$ tuples for the target task $t$, which may come either from online interventions or small set ($\sim10$) of demonstrations.
    \item $\dpr$: which contains an unsegmented mix of data from tasks in the space of $\mathcal{T}$, which may include everything from random data to near-expert data for task $t$ to sub-optimal data for any task other than $t$.
\end{enumerate} 
In practice, we expect the prior data to have some data that is relevant to task $t$ and some that is not relevant. 
Given these data sources, the final goal is to learn a policy $\pi_t: \mathcal{S} \rightarrow \mathcal{A}$ that is performant on task $t$, and to do so in a way that leverages the prior data to get better performance. In the extreme case where none of the prior data is relevant, the optimal method should ideally not be harmed by the prior data and learn a policy as good purely learning form $\dt$.

\subsection{Overview} 

In \algo, we explore going beyond the standard approach of pre-training a policy on the prior dataset $\dpr$ and fine-tuning on the task-specific dataset $\dt$. 
Our key insight is that $\dt$ not only provides data to train on but also informs what data from $\dpr$ is relevant for learning. Moreover, we can learn the metric for measuring this relevance from $\dpr$. Concretely, our method, \algo,
first learns a $(s, a)$-similarity metric from $\dpr$, then given $\dt$ uses the similarity metric to retrieve relevant $(s,a)$ tuples from $\dpr$, and finally trains $\pi_t$ on the union of $\dt$ and the retrieved data with imitation learning (See Figure~\ref{methodoverview}). We cover the similarity metric training in Section~\ref{method:similarity}, the filtering process in Section~\ref{method:filtering}, and the policy training in Section~\ref{method:training}. See Algorithm~\ref{alg} for a summary of the \algo~method.

\subsection{Similarity Metric Pre-training}
\label{method:similarity}

To effectively retrieve behaviors from $\dpr$ relevant to $\dt$, we need some way of measuring the similarity between two $(s,a)$ tuples, a non-trivial problem for high-dimensional state spaces like images. Critically, because $\dpr$ may have behaviors from completely different tasks than the target task, its not sufficient to simply pull transitions with similar states, as the actions may be poorly-suited to the target task.
Therefore, we want to learn a function $\mathcal{F}: \mathcal{S} \times \mathcal{A} \times \mathcal{S} \times \mathcal{A} \rightarrow \mathbb{R}$ which takes as input two state-action pairs and predicts a scalar score measuring their functional similarity.
The first component of our method is learning $\mathcal{F}$ using $\dpr$.

\begin{figure}[t]
    \centering
    \includegraphics[width=0.99\linewidth]{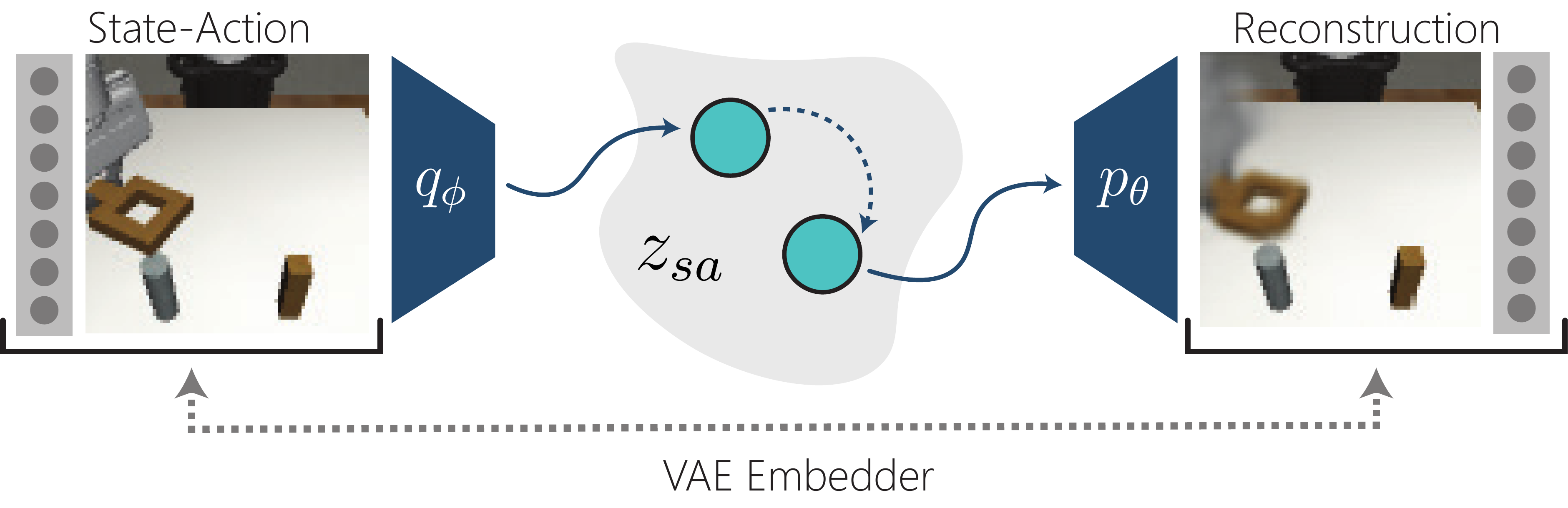}
    \vspace{-0.3cm}
    \caption{\small \textbf{Training the State-Action Embedding}. We train a variational auto-encoder jointly on states and actions to produce our state-action embedding $z_{sa}$. 
    }
    \vspace{-0.3cm}
    \label{fig:similarity}
\end{figure}

At a high level, we do so by learning a low dimensional $(s,a)$ embedding using $\dpr$, and instantiate $\mathcal{F}$ as the distance between the embeddings of $(s_1,a_1) \sim \dpr$  and $(s_2,a_2) \sim \dt$ tuples. 
There are many possible approaches to learn these $(s,a)$ embeddings, ranging from contrastive learning to compression via an auto-encoder. In our experiments we find that compressing states and actions with a VAE provides a simple approach with the best performance (See Figure~\ref{fig:similarity}).

Concretely, we simply learn a joint variational autoencoder for the states and actions, which are trained to maximize the evidence lower bound (ELBO) 
\begin{equation}
\label{eq:elbo}
    \mathbb{E}_{q_{\phi}(z_{sa}|s,a)}[\log p_\theta(s,a | z_{sa})] - \beta \mathcal{D}_{KL} [q_{\phi}(z_{sa}|s,a)|| p(z_{sa})]
\end{equation}
on state-action pairs sampled from $\dpr$, where $\beta$ is a tuned hyperparameter. The state and action are encoded separately then combined through a multi-layer perception (MLP)
to form the state-action embedding. The action is additionally concatenated to this state-action embedding as $z = [z_{sa}, a]$,
which we found to help with behavior separability.

Finally given the two tuples $(s_1, a_1)$ and $(s_2, a_2)$ we produce $z_1$ and $z_2$. The distance function, $\mathcal{F}$, is then simply the negative L2 distance between the two embeddings:
\begin{equation}
    \mathcal{F}(s_1, a_1, s_2, a_2) = -||z_1 - z_2 ||_2
\end{equation}

\begin{figure}[t]
    \centering
    \includegraphics[width=0.9\linewidth]{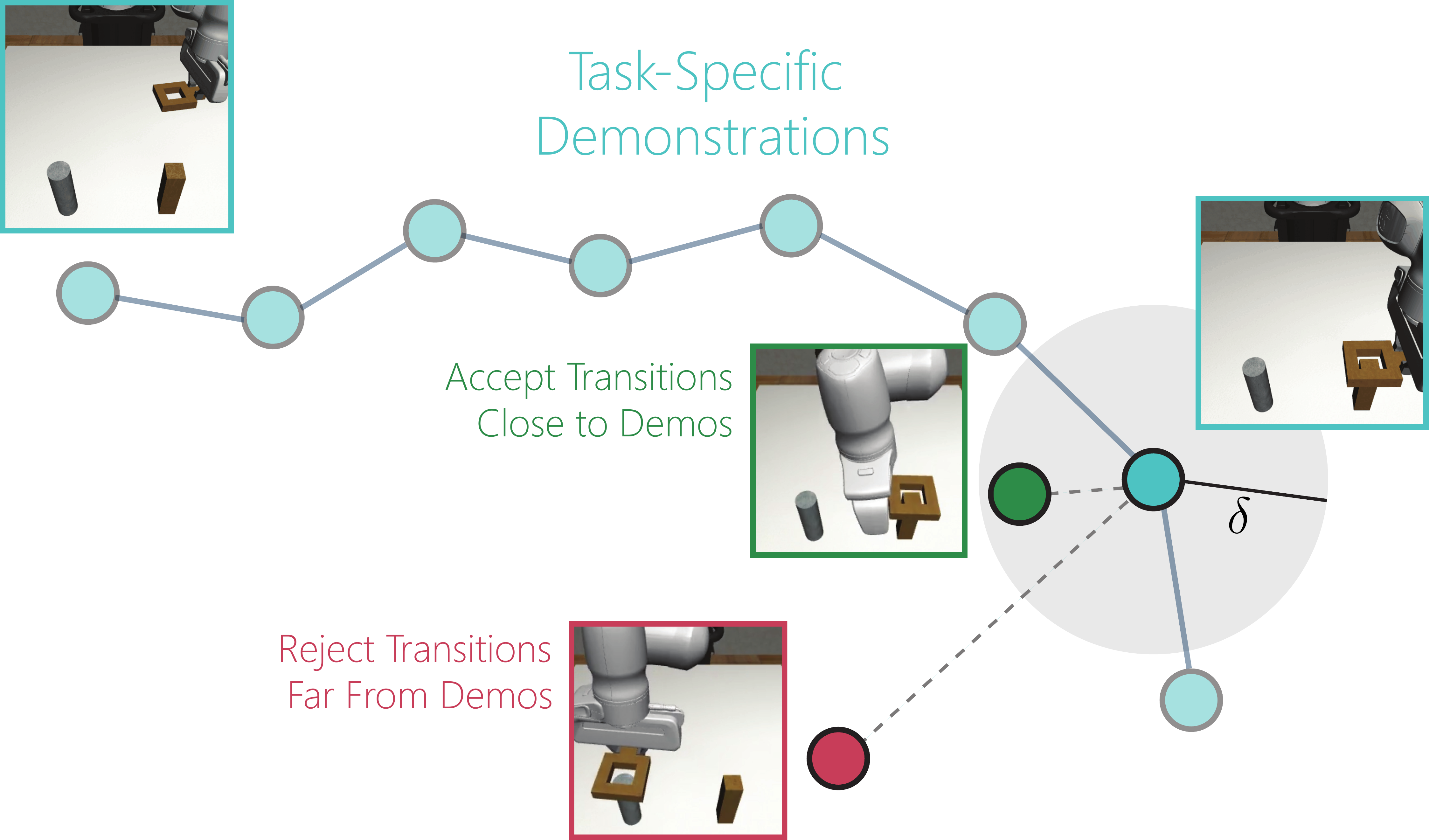}
    \caption{\small \textbf{Retrieving from the Unlabeled Dataset}. Using our pretrained state-action embedder, we compute the embeddings for the offline dataset $\dpr$ and the small number of task-specific demos $\dt$. Then, we select transitions in the offline dataset within a certain distance of the task-specific demos, in embedding space. 
    }
    \vspace{-0.3cm}
    \label{fig:filtering}
\end{figure}

\subsection{Filtering Relevant Data}
\label{method:filtering}

Given the pre-trained $\mathcal{F}$, how might we retrieve relevant data from $\dpr$ using $\dt$? While there are many valid strategies from upweighing losses to importance sampling, we hypothesize that transitions in $\dpr$ where the expert is taking similar actions in similar states will be useful to imitate for the target task, and thus should be leveraged. Therefore,
if a transition in $\dpr$ is relevant to \textbf{any} transition in $\dt$, we choose to retrieve it and learn from it with behavior cloning. 

Concretely, let the upper bound on similarity $\mathcal{F}^+ = \max_{(s^*,a^*) \in \dpr}\max_{(s,a) \in \dt} \mathcal{F}(s^*, a^*, s, a)$ and the lower bound $\mathcal{F}^- = \min_{(s^*,a^*) \in \dpr}\max_{(s,a) \in \dt} \mathcal{F}(s^*, a^*, s, a)$. Then, for any transition $(s^*, a^*) \in \dpr$, we retrieve it if:
\begin{equation}
\label{eq3}
    \frac{\max_{(s,a) \in \dt} \mathcal{F}(s^*, a^*, s, a) - \mathcal{F^-}}{\mathcal{F^+} - \mathcal{F^-}} > \delta 
\end{equation}

We take this minmax normalization across $\mathcal{F}$ to account for varying magnitudes of embeddings. 
In our experiments, we use $\delta \in [0.6, 0.8]$, although we show in section \ref{exp:e3} that our method is robust to changes in $\delta$. Since we retrieve the top $1 - \delta \%$ of $\dpr$ that we think is most relevant, when using \algo, a user should choose $\delta$ to be one minus the percent of $\dpr$ likely to be relevant. For instance, if almost none of $\dpr$ is expected to be relevant, $\delta$ should be set close to 1, and if all of $\dpr$ is relevant $\delta$ should be set close to 0. See Figure~\ref{fig:filtering} for  visual illustration of the filtering process.

\subsection{Training with Retrieved Data}
\label{method:training}

Finally, given the retrieved data, $\dr$, we then train the agent to jointly imitate $\dr$ and $\dt$ with a behavior cloning loss. That is, we train $\pi_{\psi}$ with the objective:
\begin{equation}
\label{eq4}
\begin{aligned}
    \min_{\psi} 
    \mathbb{E}_{(s, a) \sim \dt}[ - \log \pi_\psi(a | s)] +  \\
    \mathbb{E}_{(s,a) \sim \dr}[ - \log \pi_\psi(a | s)]
\end{aligned}
\end{equation}

We encode the observation modalities separately before combining them into a shared state embedding for the policy. For the policy, we use an LSTM \cite{hochreiter1997long} architecture,
which allows for object permanence amidst occlusion and the observation of features like object velocity.
The policy outputs the parameters to a Gaussian 
mixture model (GMM), and the final actions are chosen by sampling from the GMM, which allows for better modeling of multi-modal behavior often present in a large multi-task datasets. See Appendix~\ref{app:modeldetails} for more details on the policy architecture and training. Algorithm~\ref{alg} summarizes the method, with lines 5-7 describing the embedding pre-training, lines 9-11 summarizing the filtering process, and lines 13-16 describing the policy training phase.

\begin{algorithm}[t]
\caption{\algocap}
\begin{algorithmic}[1]
\label{alg}
\STATE \textbf{Input}: Prior dataset $\dpr = [(s_1, a_1), ..., (s_T, a_T)]_{1:N}$
\STATE \textbf{Input}: Task dataset $\dt = [(s_1, a_1), ..., (s_T, a_T)]_{1:10}$
\STATE Initialize $\psi$, $\phi$, $\theta$ randomly, $\dr \leftarrow \emptyset$. Batch size $B$.
\STATE \textcolor{blue}{/* Train VAE embedding space */}
\WHILE{VAE not converged}
        \STATE $s_t, a_t \sim \dpr$
        \STATE Update $\theta$, $\phi$ to maximize Eq.~\ref{eq:elbo}
\ENDWHILE
\STATE \textcolor{blue}{/*Retrieve task-relevant data*/}
\FOR{$(s,a) \in \dpr$}
\IF{Eq.~\ref{eq3} is True for $(s,a)$}
\STATE $\dr \leftarrow \dr~\bigcup~(s,a)$
\ENDIF
\ENDFOR
\STATE \textcolor{blue}{/* Fit policy to task data and retrieved data */}
\WHILE{$\pi_\psi$ not converged}
\STATE $[(s, a)]_t^{1:B} \sim \dt$
\STATE $[(s, a)]_{\text{ret}}^{1:B} \sim \dr$
\STATE Update $\psi$ to maximize $\log \pi_\psi (a | s)$  \\ on $[(s, a)]_t^{1:B}$ and $[(s, a)]_{\text{ret}}^{1:B}$
\ENDWHILE
\end{algorithmic}
\end{algorithm}

\section{Experiments}
\label{sec-exp}

The main goal of our experiments is to understand the empirical performance of \algo, both in terms of how it works and how well it works. 
After describing the experimental set-up (Section~\ref{exp:envs}), we will first examine what data is retrieved by our method in both simulated and real robotic manipulation tasks (Section~\ref{exp:e1}). We will then analyze the performance of \algo~in comparison to pre-training with finetuning (Section~\ref{exp:e2.1}), in comparison to other retrieval strategies (Section~\ref{exp:e2.2}), and in evaluations on a physical robot (Section~\ref{exp:e2.3}) using a heterogenous large-scale dataset collected in prior work \cite{Ebert2021BridgeDB}. Beyond final performance, we also study the robustness of \algo{} (Section~\ref{exp:e3}) and ablate design choices and hyperparameters (Section~\ref{exp:e4}). In the appendix, we additionally include experiments that study \algo's ability to stitch together behaviors from short-horizon tasks into a sequential task (Appendix~\ref{exp:e5}), its ability to be combined with other policy pretraining methods \cite{Lynch2019LearningLP} (Appendix~\ref{exp:e7}), its robustness to larger and more diverse $\dpr$ (Appendix~\ref{exp:e8}),  and an analysis of failure modes of \algo~and comparisons (Appendix~\ref{exp:e6}).

\subsection{Experimental Domains}
\label{exp:envs}

\begin{figure}[t]
    \centering
    \includegraphics[width=0.97\linewidth]{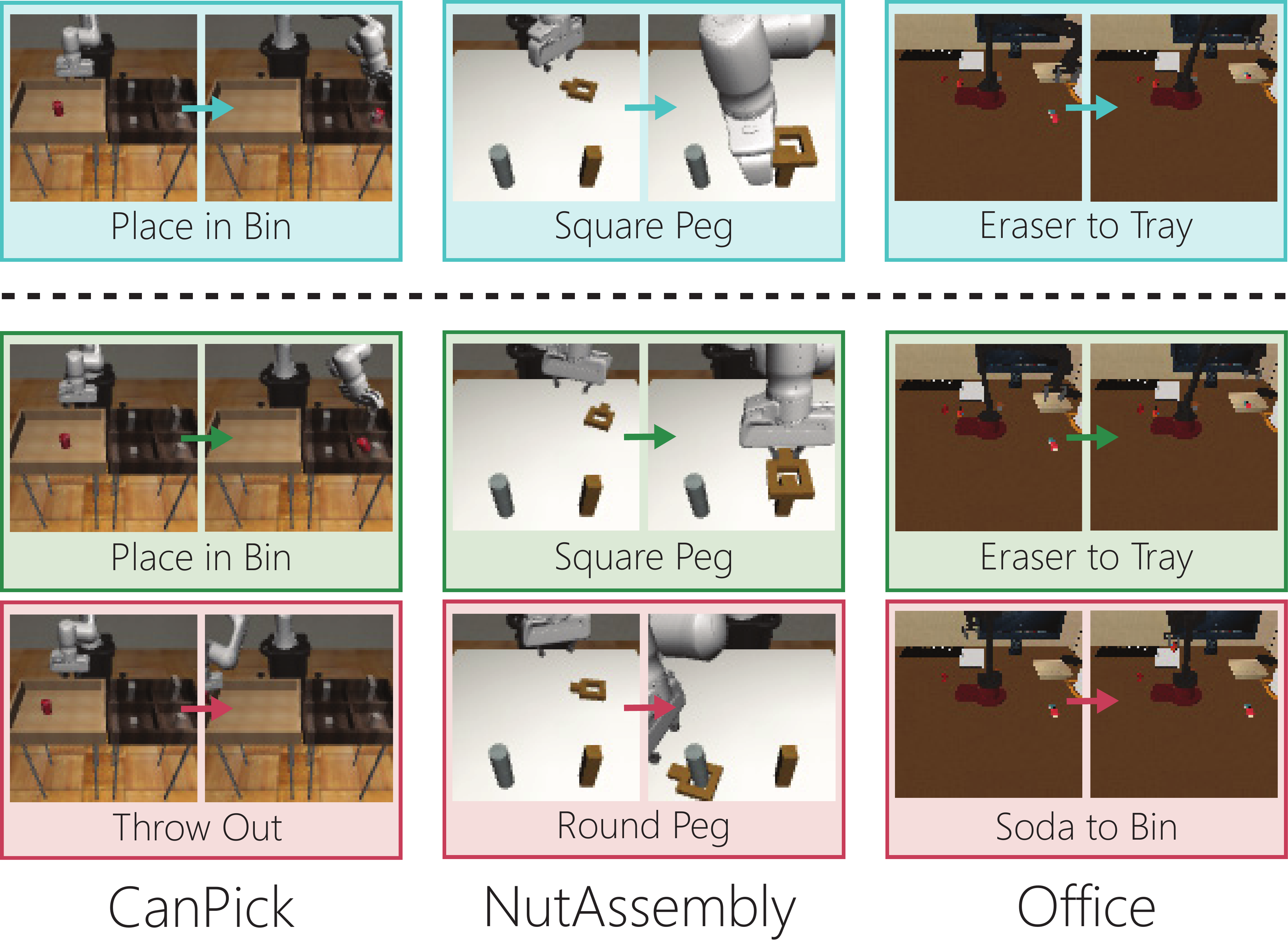}
    \vspace{-0.1cm}
    \caption{\small \textbf{Simulation Environments}. We consider three simulated domains.In each domain we highlight the downstream task data $\dt$ in \textcolor{myblue}{blue}, relevant offline data from $\dpr$ in \textcolor{mygreen}{green}, and irrelevant downstream data from $\dpr$ in \textcolor{myred}{red}. In RoboSuite Can Pick and Place (\textbf{left}), the agent must pick an place a can into the bin, and irrelecant data involves throwing the can randomly. In RoboSuite Nut Assembly (\textbf{middle}), the agent must insert a square into the correct peg, and irrelevant data involves putting it onto the wrong peg. In PyBullet WidowX Office Cleanup (\textbf{right}), the agent must pick and place an eraser into a specified tray, where irrelevant data involved many actions with other objects in the scene.
    }
    \vspace{-0.3cm}
    \label{fig:environments}
\end{figure}

\begin{figure}[t]
    \centering
    \includegraphics[width=0.97\linewidth]{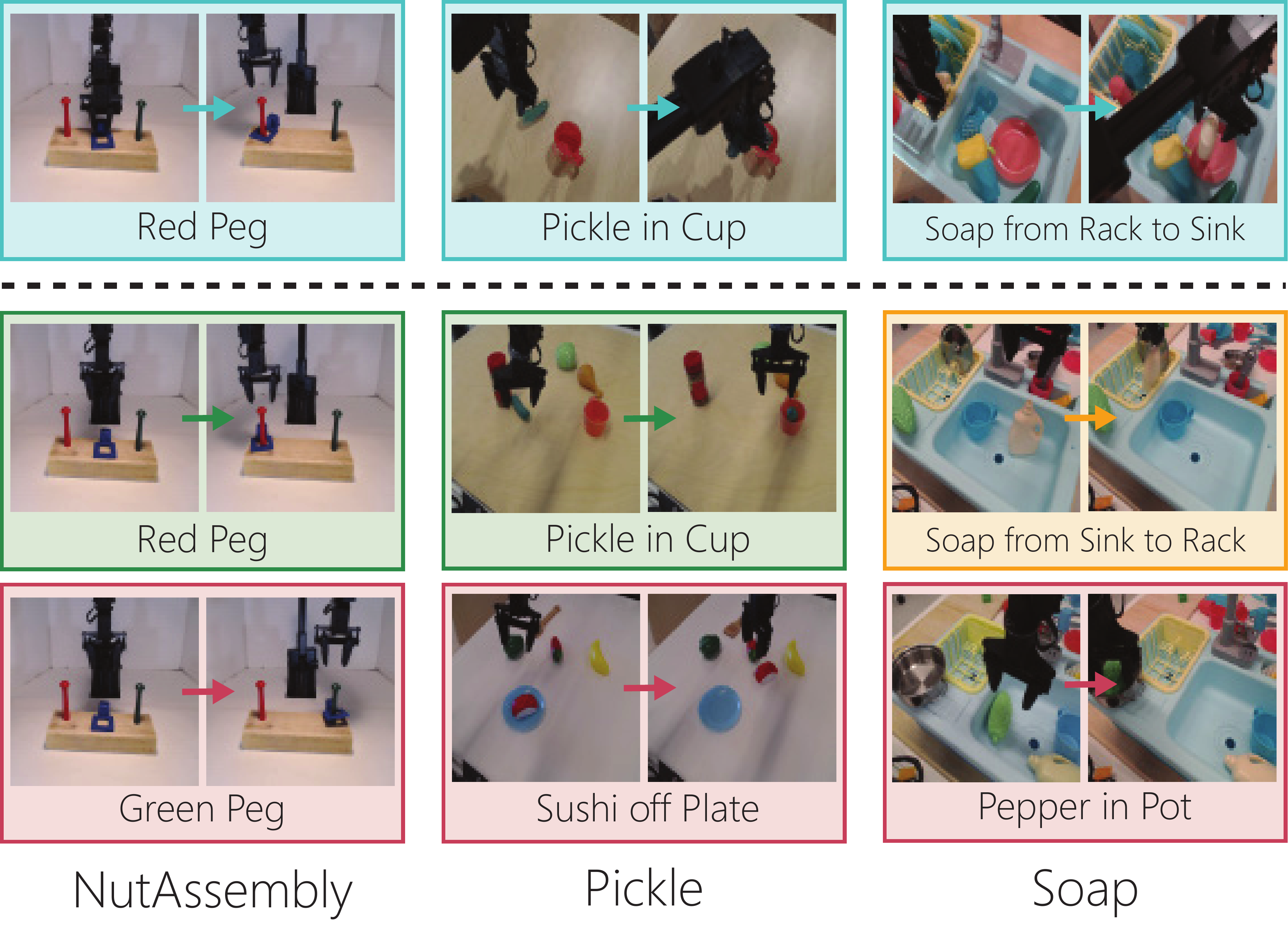}
    \caption{\small \textbf{WidowX Robot Environments}.We consider three robot domains. In each domain we highlight the downstream task data $\dt$ in \textcolor{myblue}{blue}, relevant offline data from $\dpr$ in \textcolor{mygreen}{green}, partially relevant offline data from $\dpr$ in \textcolor{myorange}{orange}, and irrelevant offline data from $\dpr$ in \textcolor{myred}{red}. In Nut Assembly (\textbf{left}), the agent must insert a square into the correct peg, and irrelevant data involves putting it onto the wrong peg (like the task in simulation). In Pickle in Cup (\textbf{middle}), the agent must pick and place a pickle in a cup, and $\dpr$ is a large subset of the Bridge Dataset collected in a similar tabletop setting\cite{Ebert2021BridgeDB}. In Soap in Sink (\textbf{right}), the agent must pick and place a soap in sink, a totally novel task, and again $\dpr$ is a large subset of the Bridge Dataset collected in a similar sink setting. \cite{Ebert2021BridgeDB}.
    }
    \vspace{-0.3cm}
    \label{fig:robot_environments}
\end{figure}

We study \algo~and comparisons in three simulated domains with 7DOF end-effector control and three real robot domains with 4DOF end-effector control, visualized in Figures \ref{fig:environments} and \ref{fig:robot_environments} respectively. The task data $\dt$ for all environments contain $10$ expert demos for the target task. 

\noindent \textbf{Sim: CanPick.}
In this RoboSuite environment \cite{robosuite2020, robomimic2021}, the task is for a simulated Franka robot to pick and place a coke can from one bin to another. $\dpr$ contains a mix of $400$ human-collected demos \cite{robomimic2021} where half complete the task and half fail by randomly throwing the can out of the bin. 

\noindent \textbf{Sim: NutAssembly.}
The modified RoboSuite \cite{robosuite2020, robomimic2021} task is for a simulated Franka robot to pick and insert a square into the right peg. $\dpr$ contains a mix of $400$ machine-generated demos where half complete the task and, in the other half, the robot puts the square onto the wrong (left) peg. 

\noindent \textbf{Sim: Office}.
The third task is for a simulated Widowx robot in an office environment \cite{singh2021parrot} to pick an eraser and place it into a specified tray. $\dpr$ contains $1200$ machine-generated demos of $8$ separate pick-place behaviors. Out of these, $\dpr$ contains $150$ demos of the target eraser-tray task.

\noindent \textbf{Real: NutAssembly}
Analogous to the simulated nut assembly task, a real WidowX robot needs to pick up a blue square hole and insert it onto the red peg. As constructed, the hole and peg have a narrow tolerance, requiring precise control for successful insertion. $\dpr$ contains a mix of $160$ human-collected demos where half complete the task and half fail by inserting the peg onto the green peg. 

\noindent \textbf{Real: Pickle}
The second real-world task is for a real WidowX robot to pick up a pickle from the table and place it into a cup randomly placed within the robot's visual field. $\dpr$ is a subset of the Bridge Data \cite{Ebert2021BridgeDB} that contains tabletop manipulations, totalling to $285$ demos. Out of these demos, $45$ correspond to an analogous pickle-cup task, although there are significant shifts in lighting, table color, and distractor objects.

\noindent \textbf{Real: Soap}
The final task is for a real Widowx robot to pick up a detergent container from the dish rack of a toy kitchen sink and place it, standing up, on a red plate in the sink. $\dpr$ is a subset of the Bridge Dataset \cite{ebert2018visual} that contains all the $485$ trajectories collected on a copy of this sink. 
There is no analogous task in the Bridge Dataset, although there are trajectories involving the detergent bottle, as well as sub-trajectories that pull other objects from the dish rack or place other objects into the sink. 

\subsection{What Data Does \algocap~Retrieve?}
\label{exp:e1}

\begin{figure}[t]
    \centering
    \includegraphics[width=0.99\linewidth]{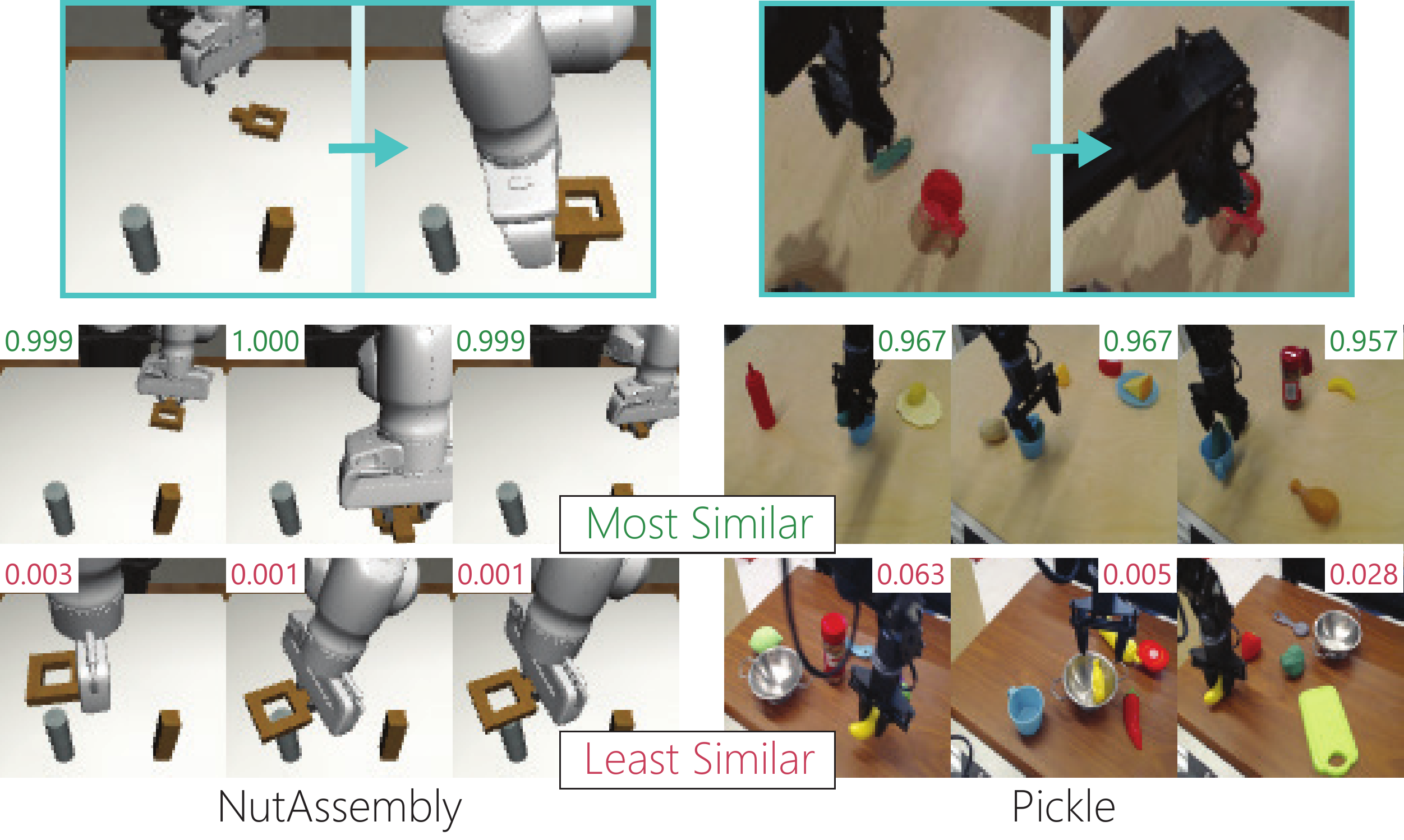}
    \vspace{-0.3cm}
    \label{fig:top_square}
    \caption{\small \textbf{Visualizing the Retrieved Data}: For each target task in $\dt$ (top), we highlight the 3 random states and their scores from the top-25 highest ranked transition by \algo~(middle), as well as 3 random states and their scores from the lowest 25 ranked transitioned by \algo~(bottom). We see that \algo~is indeed ranking highly transitions that visually look relevant to the target task in $\dt$, and rejecting states that are clearly from other tasks.} 
    \vspace{-0.3cm}
\label{fig:visualizing_selected}
\end{figure}

In our first experiment, we qualitatively analyze what type of behaviors are retrieved in the Sim: NutAssembly and Real: Pickle tasks.
In Figure~\ref{fig:visualizing_selected}, we show example states from $\dt$, and three randomly sampled states from the top-25 highest ranked transitions in $\dpr$ according to \algo~and three randomly sampled states from the lowest 25 transitions in $\dpr$ according to \algo.
We observe that indeed the states ranked highly by \algo~are relevant to the target task in $\dt$, and the lowest ranked transitions are visibly from different or adversarial tasks (e.g. moving the nut to the wrong peg).

\begin{figure}[t]
    \centering
    \includegraphics[width=0.99\linewidth]{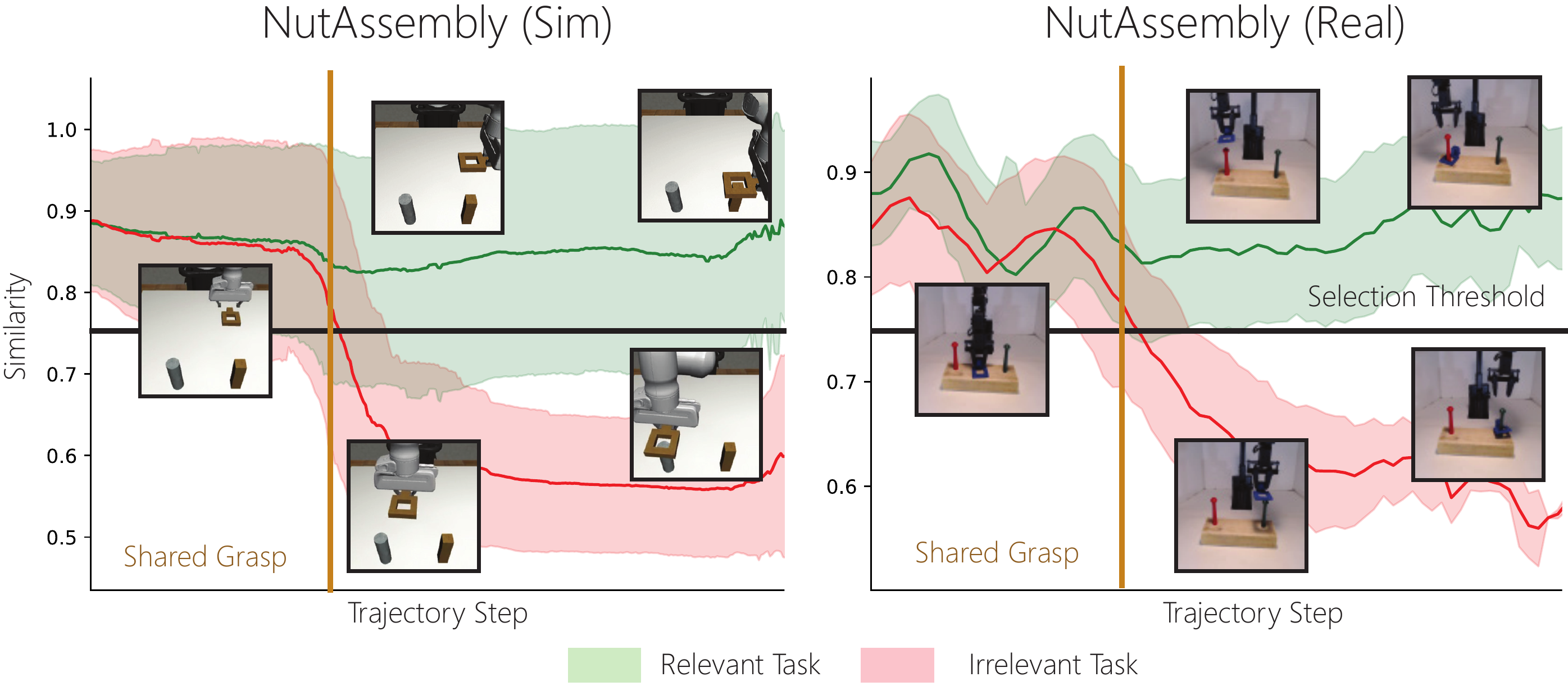}
    \vspace{-0.3cm}
    \caption{\small \textbf{Demonstration of Embedding Separation}. We measure how \algo~scores data from $\dpr$ for the target task. Green is the ground truth relevant data, while red is the ground truth irrelevant demos. We see that \algo~ learns to make use of the part of the irrelevant demos that are useful to the target task (grasping), but only highly scores the relevant demos afterwards. This highlights that \algo~is able to make use of all relevant transitions, from both relevant and irrelevant tasks, but effectively discards transitions that aren't useful to the target task.
    }
    \vspace{-0.3cm}
    \label{separationfig}
\end{figure}

In our next experiment, we quantitatively evaluate how well \algo~separates task relevant and irrelevant data, particularly looking at the NutAssembly task in both simulation and the real robot (other environments in the Appendix~\ref{app:viz}).
Concretely, we plot the average embedding similarity for the ground truth relevant data (green) and for the ground truth irrelevant data (red) (See Figure~\ref{separationfig}). For example, in the nut assembly task data from $\dpr$ that puts the square on the correct peg is green, and on the incorrect peg is red. While the agent is grasping the square, \algo~retrieves data from both sources as the grasp is always relevant to the downstream task. However, once the agent has grasped, we see the score for the task-irrelevant data drop significantly. This highlights \algo's ability to make use of all transitions in $\dpr$ that are useful for the target task, while effectively filtering out those that are not useful, a valuable property for being able to leverage large, heterogeneous datasets as $\dpr$.

\begin{figure*}[t]
    \centering
    \includegraphics[width=0.99\linewidth]{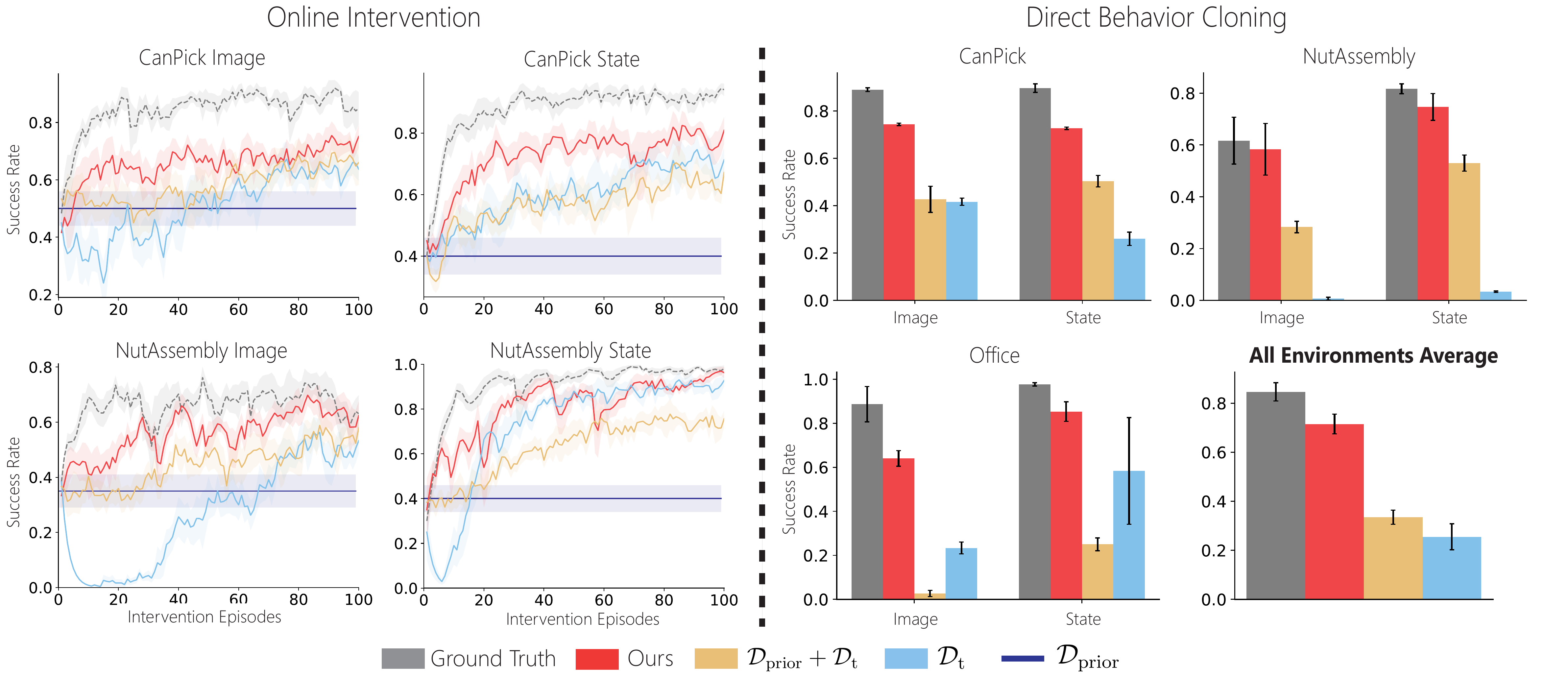}
    \caption{\small \textbf{Comparing \algocap~to Other Retrieval Strategies}. In simulation, we compare the success rate of \algo~(\textbf{Ours}) to other strategies for retrieval including only training on $\dt$ \textbf{($\dt$)}, balanced batches of $\dpr$ and $\dt$ \textbf{($\dpr + \dt$)}, only \textbf{($\dpr$)} (interventions only), and using \textbf{Ground Truth} separation which uses privileged information. We consider both online interventions (top) and 10 demos (bottom) as $\dt$. In both cases we see \algo~approach the ground truth oracle performance, while significantly outperforming the comparisons, which either overfit to only using $\dt$, or train too much on sub-optimal data from $\dpr$. All results are computed over 100 trials (or 20 trials every episode in the intervention setting), all averaged over 3 seeds with standard error shading.}
    \vspace{-0.3cm}
    \label{Exp2results}
\end{figure*}

\subsection{Does \algocap~Improve Performance ...}
\label{exp:e2}

\subsubsection{Over Other Pre-training+Finetuning Schemes?}
\label{exp:e2.1}

First, we aim to understand how \algo~compares to other techniques that use pre-training and adapt the policy on a target dataset. A conventional approach is to perform multi-task pre-training of $\dpr$, followed by fine-tuning on the target task data $\dt$.
We begin by comparing \algo~\textbf{(Ours)} to such methods in simulation: pre-training with goal-conditioned behavior cloning on $\dpr$ \textbf{(GCBC $\dpr$)}, as well as with additional finetuning of the goal-conditioned policy on balanced batches of $\dt$ and $\dpr$ \textbf{(GCBC $\dpr$ +FT $\dt$)}. See Appendix~\ref{app:modeldetails} for more details on the implementation of GC and GC+Finetune.
\begin{table}[t]
  \centering %
  \begin{tabular}{c|ccc} 
    \toprule     
    &  GCBC $\dpr$ & GCBC $\dpr$ & Ours \\ & & + FT $\dt$ &  \\
    \midrule
    \vspace{0.05cm}
    CanPick (Image)  & \textbf{$68 \pm 14\%$} & $62 \pm 2\%$& \textbf{74 $\pm$ 1\%}  \\
    \vspace{0.05cm} CanPick (State) &  $33\pm 1\%$ & $45 \pm 1\%$ &  \textbf{73 $\pm$ 1\%}\\
     \vspace{0.05cm}NutAssembly (Image)  & $22 \pm 4\%$ & $10 \pm 1\%$ & \textbf{58 $\pm$ 9\%} \\
    \vspace{0.05cm} NutAssembly (State)  & $15 \pm 1\%$ & $10 \pm 1\%$ & \textbf{75 $\pm$ 5\%} \\
    \vspace{0.05cm} Office (Image)  & $0 \pm 0\%$ & 10 $\pm$ 4\%& \textbf{64 $\pm$ 3\%}  \\
    \vspace{0.05cm} Office (State)  & $76 \pm 5\%$ & $76 \pm 1\%$ & \textbf{85 $\pm$ 4\%} \\
    \midrule
    \textbf{Average} & 35 $\pm$ 4\% & 36 $\pm$ 2\% & \textbf{72 $\pm$ 2\%} \\
    \bottomrule
  \end{tabular}
  \caption{\small \textbf{Comparing \algocap~to Pre-training + Finetuning in Simulation}.  We report the success rate on our three simulated tasks across 100 trials (averaged over 3 random seeds with standard error shown) of \algo~compared to goal-conditioned behavior cloning (GCBC) on $\dpr$ and with finetuning on $\dt$. We see that both with and without fine-tuning multi-task pre-training struggles and \algo~performs best across all settings, improving performance by over 30\% on average.}
  \label{tab1}
  \vspace{-0.3cm}
\end{table}
In Table~\ref{tab1} we observe that across all three simulated tasks,
\algo~improves over goal-conditioned BC, with and without finetuning, improving the success rate by 30\% on average. 

We hypothesize that purely offline goal-conditioned BC struggles with underfitting, particularly in cases where goal-states may may over-specify the task (i.e. many goal-states can represent the same task). Similarly, adding fine-tuning can be unstable -- for example if training only on $\dt$ the fine-tuned policy can overfit, while if training on balanced batches with the offline data, the underfitting issue remains. This is particularly the case in settings where the desired change in state only makes up a small portion of the goal-states, like in Office (Image). Ultimately \algo~is able to pull only the relevant data from $\dpr$, and train on all of it without overfitting to $\dt$, resulting in the best performance.

\begin{figure*}[t]
    \centering
    \includegraphics[width=0.9\linewidth]{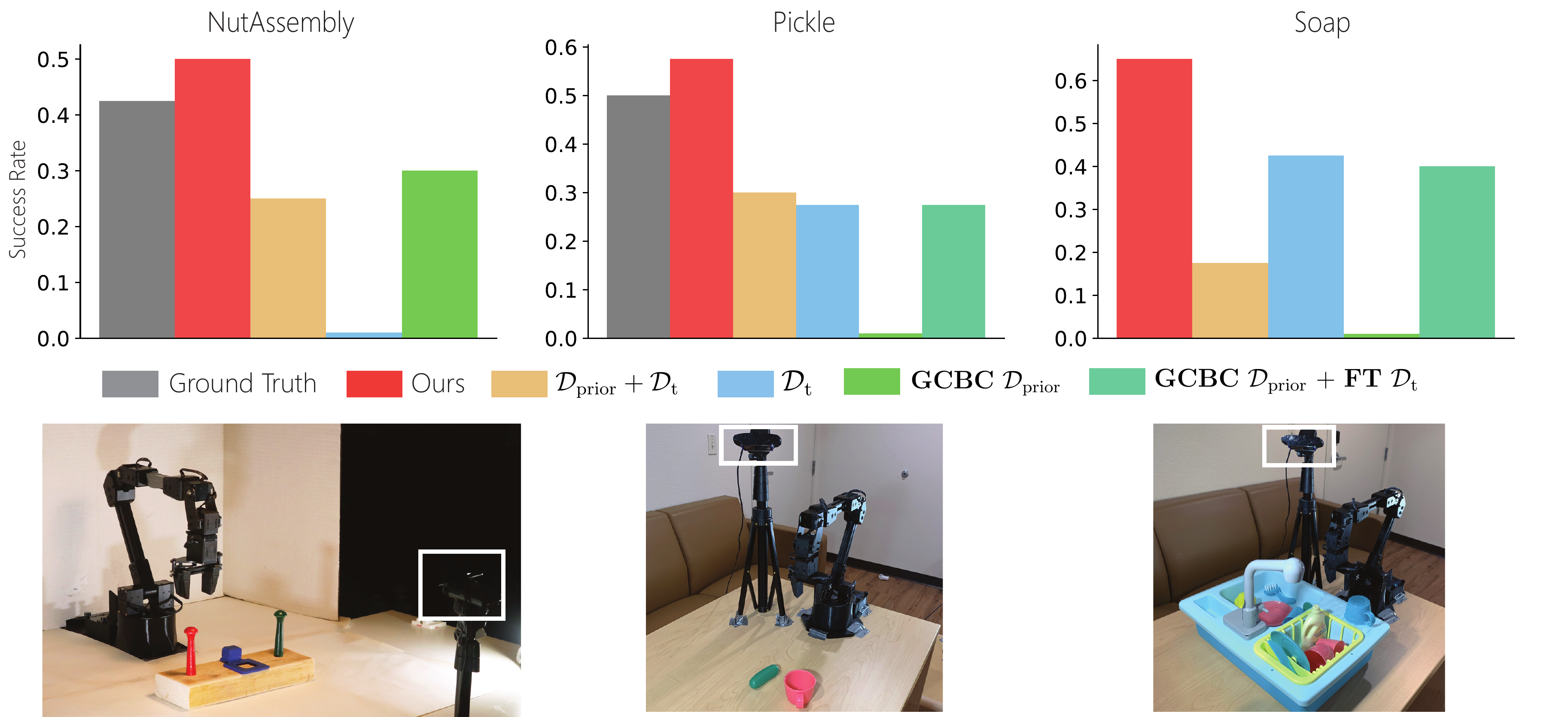}
    \caption{\small \textbf{Real Robot Comparisons}. Across our three robot tasks, we compare the success rate given 10 target demos across  40 evaluations of \algo, other retrieval methods, and pre-training and fine-tuning (\textbf{top}). We again see that \algo~improves over the comparisons by over 20\% in success rate. Because it is able to leverage the useful transitions even from irrelevant tasks in the offline data, \algo~exceeds using the ground truth filtering which uses only data from $\dpr$ from the same exact task. We show the robot setup for each task (\textbf{bottom}), as well as the camera capturing observations highlighted in the white box.
    }
    \vspace{-0.3cm}
    \label{robotexps}
\end{figure*}

\subsubsection{Over Other Retrieval Strategies?}
\label{exp:e2.2}

We also compare \algo~\textbf{(Ours)} to other retrieval strategies across all of our simulated environments. This includes not doing any retrieval \textbf{($\dt$)}, training on balanced batches from $\dpr$ and $\dt$ \textbf{($\dpr + \dt$)}, and using the ground-truth task label to filter only task-relevant demos from $\dpr$ \textbf{(Ground Truth)}. This last approach can be thought of as a privileged Oracle, that does not use our learned filtering metric, but instead uses the ground truth label to filter. This will retrieve demos from $\dpr$ that are known to be the same task, however will not retrieve transitions from a different task that may be relevant. 
We consider both the setting where downstream expert data comes in the form of online interventions (Figure~\ref{Exp2results} (\textbf{left})) and when it comes in the form of 10 demos (Figure~\ref{Exp2results} (\textbf{right})).

We observe in Figure~\ref{Exp2results} that despite the expert data coming in very different forms, whole demos vs. online interventions, in both settings \algo~significantly improves success rate over other retrieval strategies, by over 20\% given a small set of demos, and by around 10\% given online interventions. In the online intervention setting, the human supervision is strong (on-policy corrections), therefore we expect $\dt$ to particularly useful and using $\dpr$ to be less important, hence why we see smaller gains from \algo. In the setting where the agent is given a small set of demonstrations (Figure~\ref{Exp2results} (\textbf{right})), \algo~nearly matches using the oracle retrieval from $\dpr$, suggesting that it is effectively capturing the relevant data to $\dt$.

In the case of online interventions, $\dt$
struggles to learn from the small amount of online expert data, and thus can overfit significantly or experience sharp drops in performance (Figure~\ref{Exp2results} (\textbf{left}), NutAssembly and CanPick (Image)). Even in the case of a handful of target demos, $\dt$ has larger variance (Figure~\ref{Exp2results} (\textbf{right}), Office (State)). 
Na\"ively using balanced batches of $\dpr$ and $\dt$ can lead to training with task-irrelevant data, leading to worse task-specific performance. 

Comparing online interventions overall (Figure~\ref{Exp2results} (\textbf{left})) to full expert demos (Figure~\ref{Exp2results} (\textbf{right})) for $\dt$, we observe that on average performance with online expert data is slightly better, an expected result as the human supervision is in-the-loop and on the states the agent most needs feedback. Additionally, we see on average that learning from low-dim state information tends to perform slightly better than from images, with some variation across tasks, again matching our intuition that lower-dimensional state space of object and robot positions is easier to learn from.

\subsubsection{On a Real Robot?}
\label{exp:e2.3}

While \algo~performs effectively, do the benefits translate to a physical robot? In this experiment, we compare \algo~to all of the above comparisons from Sections~\ref{exp:e2.1} and ~\ref{exp:e2.2} on a real WidowX robot. We test on three real robot tasks, NutAssembly, Pickle, and Soap, as described in Section~\ref{exp:envs}, where $\dpr$ for the Pickle and Soap tasks involve using a subset the pre-existing Bridge Dataset \cite{Ebert2021BridgeDB}.

In our results (Figure~\ref{robotexps}), we again see that \algo~improves over both other retrieval strategies and pre-training + finetuning by more than 20\% in success rate. In fact, we see that because it is able to use useful transitions even from irrelevant tasks in the offline data, \algo~exceeds using only prior data from the same task (Ground Truth),  which uses privileged information in the form of task labels.

Additionally, we note that like in our simulation experiments, only using $\dt$ tends to be unstable, completely failing on some tasks like NutAssembly. Unlike our simulation experiments, we see that adding finetuning to the goal-conditioned pre-training has a much more significant impact on performance, improving over GCBC by over $30\%$. We suspect this is the case because there is a large shift from the Bridge Data $\dpr$ to the downstream task specific environment, so fine-tuning is much more important. 

\subsection{How Robust is \algocap?}
\label{exp:e3}

In our next experiment, we study how robust \algo~is with respect to distribution shifts, embedding spaces, and hyperparameters.

First on the real robot Pickle in Cup task, we study testing on a different colored cup than training, using a real pickle instead of a toy pickle, visual distractors, and physical perturbations. 
For visual distractors, we place four toy kitchen items in the visual field of the robot and shuffle them for each trial. For physical perturbations, we allow the robot to begin the alignment process with the pickle in its grasp, and then we move the cup to a different location.

\begin{table}[t]
  \centering %
  \begin{tabular}{c|cc} 
    \toprule     
     & $\dt$ & Ours\\
    \midrule
    Different Cup Color  & 1/10 & \textbf{6/10}  \\
    Real Pickle  & 0/10 & \textbf{4/10} \\
    Distractors  & 0/10 & \textbf{4/10} \\
    Physical Perturbation  & 3/10 & \textbf{4/10} \\
    \bottomrule
  \end{tabular}
  \caption{\small \textbf{Robustness to Distribution Shifts}. We test \algo's robustness to distribution shifts at test time, including changing the cup color. We find effectively using $\dpr$ to improve robustness , maintaining performance close to a 50\% success rate while using only task-specific data ($\dt$) drops considerably to around a 10\% success rate.}
  \label{robustness}
  \vspace{-0.4cm}
\end{table}

In Table~\ref{robustness} we find that across all distribution shifts, \algo~is able to maintain strong performance (nearly a 50\% success rate), while the model using only task-specific data completely collapses to around a 10\% success rate. We hypothesize that retrieving broader data that is still useful to the task makes the trained policy robust to such variation. Moreover, \algo~is able to make use of the Bridge Data \cite{Ebert2021BridgeDB}, which contains a WidowX250, while our evaluations are carried out on a WidowX200 robot, further highlighting its robustness to distribution shifts.

\subsection{Ablating Components of \algocap}
\label{exp:e4}

In this section, we ablate different design choices and hyperparameters in \algo, specifically the choice of (1) $(s,a)$ embedding space and (2) filtering threshold $\delta$.

\smallskip \textbf{Ablating $(s,a)$ Embedding.} We begin by comparing our VAE embedding space to alternative methods for learning the $(s,a)$ embedding. Concretely, we also explore using contrastive RL \cite{Eysenbach2022ContrastiveLA}, which learns an $(s,a)$ embedding that captures the future distribution of visited states with contrastive learning. In principle, this embedding should capture a functional embedding of $(s,a)$. We then explore using \algo~swapping in a contrastive RL embedding instead of our default VAE $(s,a)$ embeddings.  

\begin{table}[t]
  \centering %
  \begin{tabular}{c|cc} 
    \toprule     
    &  Ours w/ Contrastive & Ours w/ VAE (default) \\
    \midrule
    \vspace{0.05cm} CanPick (Image)  & 47 $\pm$ 1\% & \textbf{74 $\pm$ 1\%}\\
    \vspace{0.05cm} CanPick (State)  & 55 $\pm$ 3\% & \textbf{73 $\pm$ 1\%}\\
    \vspace{0.05cm} NutAssembly (Image)  & 30 $\pm$ 3\% & \textbf{58 $\pm$ 9\%}\\
    \vspace{0.05cm} NutAssembly (State)  & 19 $\pm$ 3\% & \textbf{75 $\pm$ 5\%}\\
    \vspace{0.05cm} Office (Image)  & N/A & \textbf{64 $\pm$ 3\%}  \\
    \vspace{0.05cm} Office (State)  & 39 $\pm$ 1\% & \textbf{85 $\pm$ 4\%} \\
    \midrule
    Average & $38 \pm 2\%$ & \textbf{72 $\pm$ 2\%} \\
    \bottomrule
  \end{tabular}
  \caption{\small \textbf{Comparing State-Action Embedding Spaces for Retrieval}. In our simulation tasks we compare the success rate over 100 trials (and 3 random seeds) of our method with different choices for learning the state-action embedding space. We find that using other techniques like contrastive RL to learn the state-action embedding can be unstable, leading to worse performance than the simple VAE in all tasks, and diverging in the case of Office (Image).}
  \label{tab2}
  \vspace{-0.4cm}
\end{table}

In Table~\ref{tab2}, we observe that the VAE embedding performs better across all simulated tasks. While in principle contrastive RL should capture more ``functional'' state-action embeddings, we found it to be unstable, even diverging and failing to train the representation effectively in the Office environment from image observations. 

\begin{figure}[t]
    \centering
    \includegraphics[width=0.99\linewidth]{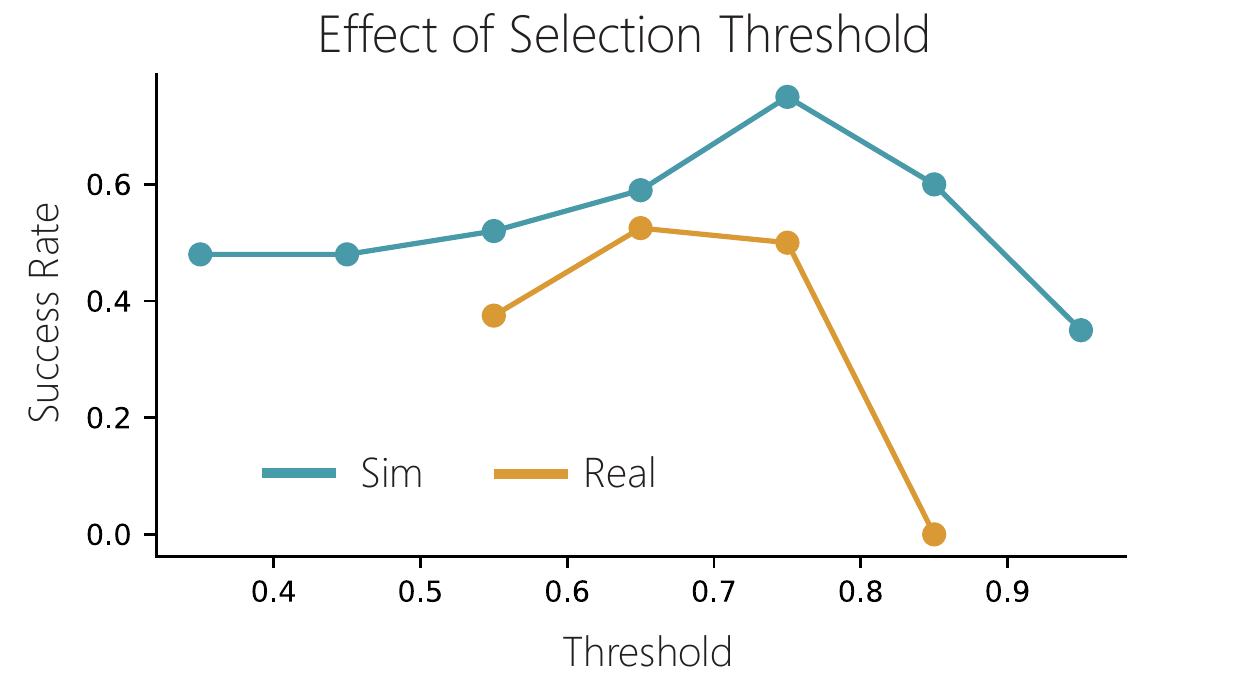}
    \vspace{-0.3cm}
    \caption{\small \textbf{Impact of Selection Threshold}. On the NutAssembly task in both simulation and the real world, we find that \algo~is generally robust to the choice of $\delta$ with the best performance in both sim and real with $\delta$ around 0.75. Interestingly, we see performance scales gracefully as $\delta$ decreases, but performance drops significantly as $\delta$ increases (filtering out all of $\dpr$).
    }
    \vspace{-0.3cm}
    \label{fig:thresholds}
\end{figure}

\smallskip \textbf{Ablating $\delta$.}
Additionally, we study the method's robustness to hyper-parameters, specifically the embedding distance threshold $\delta$ for when to accept or reject samples from $\dpr$. In Figure~\ref{fig:thresholds}, we evaluate different thresholds in simulation and on the real robot. First we see that, across domains, a threshold of around $\delta=0.75$ works best. Interestingly, we find that the method is robust to lower values of $\delta$, and performance degrades gracefully. Intuitively, this means that the method is sampling more task-irrelevant data, but is still capturing all the task-relevant data. On the other hand, performance drops sharply with too high of a $\delta$, which again makes sense: too high of a $\delta$ will throw out all of $\dpr$ including useful data.

\section{Discussion}
\smallskip \noindent \textbf{Summary.} We have presented~\algo, a method for leveraging large, unlabeled offline datasets to efficiently learn target tasks from a small amount of expert data. Our key insight is that rather than simply pre-training and finetuning, we can learn to selectively retrieve only the useful data from the prior dataset to improve task-specific training. \algo~presents a simple technique for tackling this problem, and in our experimental evaluations outperforms both multi-task pre-training and finetuning approaches as well as other retrieval strategies. Moreover, we see qualitatively that \algo~indeed retrieves task-relevant transitions while discarding those that are irrelevant.

\smallskip \noindent \textbf{Limitations and Future Work.} While we are excited about the potential of \algo, a number of important limitations remain. First, a major limitation is that our current lookup relies on similarity in a compressed embedding space of states and actions. While this has worked effectively in our experiments, it focuses on general visual similarity and may not capture the task-relevant components of a state and action in all scenarios, e.g. in cases where $\dpr$ is visually very different from $\dt$. Behavior retrieval also relies on access to the prior dataset when learning the new task, which may prevent its applicability to memory-limited scenarios where using only a pre-trained model is feasible. Finally, while we suspect that \algo~can be naturally combined with many other directions of research towards more efficient imitation learning (like pre-trained representations, restricted action spaces, and large-scale pre-training), we have yet to experimentally verify that this is the case. Combining \algo~with these methods and seeing if the benefits compound is an exciting direction for future research.

\section*{Acknowledgments}
The authors thank Jonathan Yang and Sasha Khazatsky for their assistance with the real robot setup, Suneel Belkhale for assistance in implementing the LMP baseline, and the authors of the Bridge Dataset and RoboMimic suite for their help throughout the project. This work is funded by ONR grants N00014-22-1-2293 and N00014-22-1-2621.

\bibliographystyle{plainnat}
\bibliography{ref}

\newpage

\appendices
\section{Model Details}
\label{app:modeldetails}

\subsection{Data Modalities}
We include a third-person  camera view (84 x 84 x 3), an eye-in-hand camera view (84 x 84 x 3), end-effector position (3-dim), end-effector orientation (4-dim), and gripper state (2-dim). For state-only experiments in simulation, we use the provided object states and the same proprioception.

For real robot experiments, we use the end-effector position and a third-person camera view, either mounted facing the front of the robot (nut assembly task) or looking directly down at robot's work area (tasks from the Bridge data)

\subsection{Policy Architecture}
We used a memory-augmented GMM policy. 
Images are processed with independent ResNet-18 encoders with a modified spatial softmax pooling and last layer to give a 64-vector embedding for each image. These embeddings are then concatenated with the proprioceptive lowdim data before being encoded in a two-layer LSTM with a horizon of 10 and a hidden dimension size of 1000. 

To compute the action, the LSTM-encoded state representation is passed through an MLP to yield the parameters of a GMM with 5 modes. The model is trained through log-likelihood. During inference, we sample from the GMM with a scaled-down variance. The policy is trained end-to-end, including the visual embedders. 

\subsection{VAE Embedding Architecture}
Similar to the policy modality encoder, the VAE embedder uses a ResNet-18 to encode all images, and concatenates the lowdim proprioceptive and/or object states to the image embeddings. In addition, the action is treated as a lowdim modality and concatenated to the image embeddings.

From this modality embedding, the VAE runs an MLP encoder consisting of two hidden layers (300, 400) and output a latent 128-vector. We use a fixed gaussian prior, and we use a $\beta$ weight of $0.0001$ on the KL loss. 

During training, we reconstruct the input through the VAE embedding. Low-dimensional states are recreated directly using MLPs, and the images are recreated using a ResNet-18 structure with transposed convolutions and upsampling. 

For the real robot tasks on Bridge data, we preloaded R3M  \cite{Nair2022R3MAU} weights on the ResNet-18 encoders, which were helpful in getting the model to adapt to the visually diverse data. 
\subsection{Tuning and Training}

For the VAE Images, we train for 500k steps with batch size 50. For the VAE lowdim, we train for 500k steps with batch size 100. We can monitor the progress of VAE training by examining the quality of its image reconstruction. In practice, this was sufficient to tune the main hyperparameter $\beta$, although we also made use of plots like those in Figure \ref{separationfig}. 

The plots like Figure \ref{separationfig} also allowed us to estimate the best $\delta$, although in practice, the exact $\delta$ is not critical, as long as it is not too high. 

For policies on images, we train for 300k steps with batch size 16. We use random crop augmentation on the sim environments, and we also use color jitter for the real environments. For policies on state, we train for 600k steps with batch size 100.

For our method, all data, and goal-conditioned fine-tuned, we sample from two datasets (offline, task-specific) and so the batch sizes are doubled. 

\subsection{Goal Conditioned}
We construct the goal-conditioned baseline by treating the last state in the demonstration as an input to the policy. We train only on the offline data, and during inference, we take the last state from a trajectory sampled from $\dt$, which we assume access to during test-time. 
\subsection{Goal Conditioned Fine Tune}
To fine-tune the goal-conditioned policy, we load the pretrained goal-conditioned model and then train on a balanced batch of offline and task-specific data. 

\subsection{Data Selection}
We precompute the embeddings for the offline and task dataset, and then compute the batch negative L2 distance. 

Because we use a memory-augmented policy, we implicitly assume that the past 10 steps of a selected transition are also relevant. This is not a very strong assumption, as any sequence of state-actions that get us to a task-relevant state should be considered by our task policy. 

\subsection{Our Method on Interventions}
The core of our method stays the same with the HG-DAgger experiments, except that the task-relevant data comes from the last 1000 transitions where the expert has intervened.

The expert policy is either a machine policy or a policy trained on a larger dataset of task data. Every step, we take the L2 distance of the apprentice policy action and the expert action. If the L2 distance is above a certain threshold, we take the expert action and mark the transition as an intervention. Otherwise, we take the apprentice policy action. 

We pre-train the policy with the whole offline dataset. While this yields a policy that does not know which location to place the peg or the soda can, the pre-training gave the policy the ability to grasp objects and also gave it features that made it adapt relatively quickly to the interventions. 
\section{Environment Details}
\subsection{Generating the Offline and Expert Data in Sim}
For the NutAssembly and Office simulation tasks, machine policies were used to generate the offline dataset. For NutAssembly, we use a hand-coded algorithm to grasp the Nut piece, rotate it to align the peg, and then plunge onto the peg. We add gaussian noise to the actions, and we randomly select the peg to insert onto. 

For Office, we use the provided Roboverse machine policy \cite{singh2021parrot}. We used four objects and two bins, totaling to eight separate pick-place tasks. 

\subsection{Real Robot Setup}
We used a Widowx 200 with 3DoF delta control running at 3Hz. The images come from a Logitech C920 camera secured in place during collection and testing. The testing location had consistent lighting during all trials. 

Task-relevant data and the offline data (for the nut assembly environment) were collected using a PS3 controller. For consistency, only one experienced human demonstrator was used. The 10 task-relevant data took around 10 minutes to collect. 
\subsection{Real NutAssembly Task Setup}
This setup was constructed to mimic the simulation task. Two wooden coathanger pegs were secured into a wooden base and painted red and green. The nut was constructed from MDF Plywood and painted blue. An additional piece of wood was glued to the top of the plywood to allow for easier grasping. The hole has a $\sim 4$mm tolerance on the peg, and the peg has a larger head that makes it likely that a poorly aligned hole will fall to the side. 

\subsection{Pickle Task Setup}
This setup was constructed to mimic the pickle cup task found in the bridge data. The same red cup and a similar pickle was used in the task demonstration. The red cup was randomly placed in 20 cm x 8 cm region for every demonstration and evaluation. The pickle was placed in 10 cm x 4 cm region for every demonstration and evaluation. While the bridge data contains distractors and multiple colors of cups, the task demonstrations only contained the pickle and the red cup. 

For a run to be counted as a success, the pickle must be completely inside the cup. We exclude the rare cases where the pickle rests on top of the cup. 

\subsection{Real Soap Task Setup}
This setup was constructed to mimic the sink tasks found in the bridge data. The blue sink is placed at roughly the same position as that of the Bridge Data, with a collection of toy kitchen items like cups, plates, and toy foods. The soap is placed in the corner of the dish rack, and the red plate target is placed randomly in the sink for every demonstration and evaluation. The distractor items are randomized for every trial as well. 

For a run to be counted as a success, the soap must be on the red plate and standing up by itself. In the rare cases where the plastic gets stuck to the robot gripper, we redo the trial. 

\subsection{Using the Bridge Data}
For the pickle-cup task, we selected the tabletop manipulation trajectories from the bridge data, which consisted of manipulations on three different colors of tables with various objects. For the soap in sink task, we selected all of the trajectories that are performed on this particular sink. 

For both tasks, we set up our environment to match that of the bridge data as closely as possible. This included approximating the position of the fixed camera, and we also applied an approximate transformation in the proprioceptive data to match our setup. Because we were using only 3DoF control, we discarded the orientation in the proprioception and actions of the bridge data.

\section{Additional Visualizations}
\label{app:viz}
\begin{figure}[htb]
    \centering
    \includegraphics[width=0.5\linewidth]{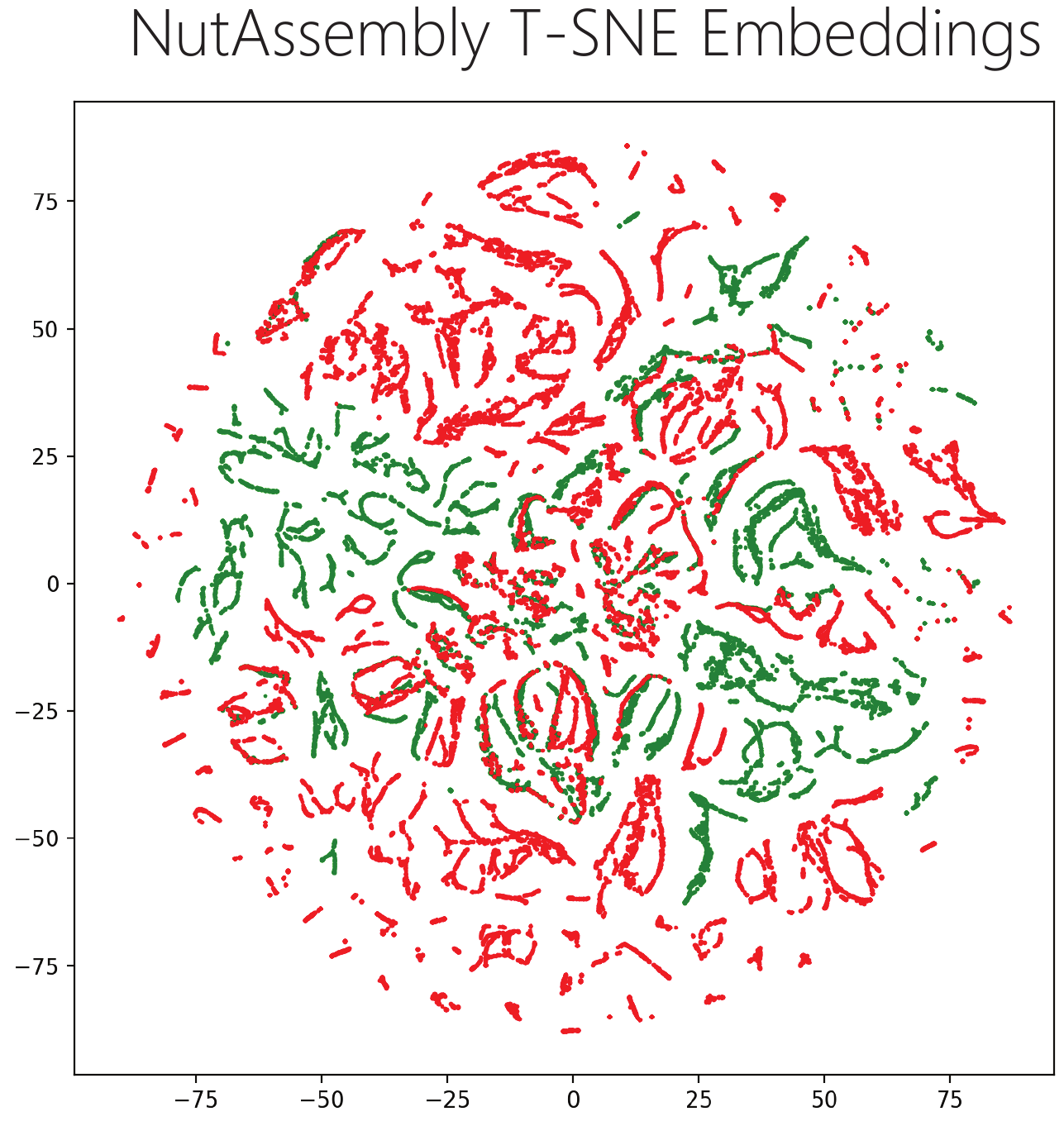}
    \caption{\small \textbf{Visualizing the Embedding Space}: Using a T-SNE projection and access to ground-truth task labels (which determine the color of the point), we are able to see how trajectories follow paths in the projection, and how the different task types find natural clusterings.}
    \vspace{-0.3cm}
    \label{app:tsne}
\end{figure}

Figure \ref{app:tsne} shows a projection of the embeddings for 100 NutAssembly simulation trajectories. We see curves in the embedding plot which consist of individual or grouped transitions. Furthermore, we see how there are distinct clusterings of trajectory types, and we also see overlaps of trajectories, which we understand to be the shared sub-trajectories. 
\begin{figure}[t]
    \centering
    \includegraphics[width=0.9\linewidth]{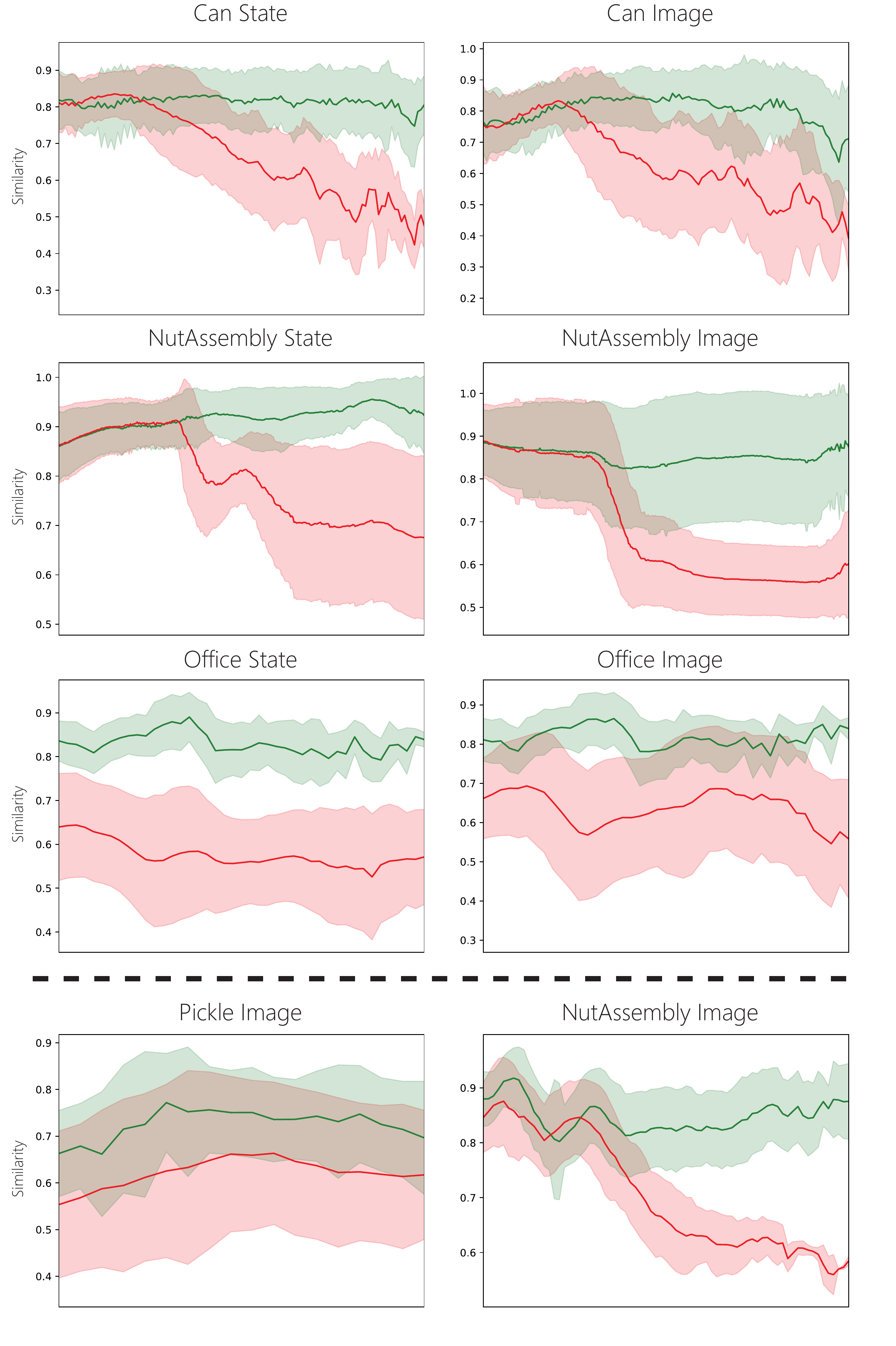}
    \vspace{-0.3cm}
    \caption{\small \textbf{Embedding Separation Plots}: We show the embedding separation plots for all of the environments with ground truth task labels.}
    \vspace{-0.3cm}
    \label{app:allseparationfig}
\end{figure}

Figure \ref{app:allseparationfig} shows all of the separation plot for the six simulation configurations and the two real robot tasks with ground truth task labels. As can be seen, our VAE embeddings create a separation between relevant and irrelevant tasks. For the CanPick and NutAssembly (both sim and real), we see an overlapping of similarities where there is a shared grasping of the target object.

\begin{figure}[t]
    \centering
    \includegraphics[width=0.99\linewidth]{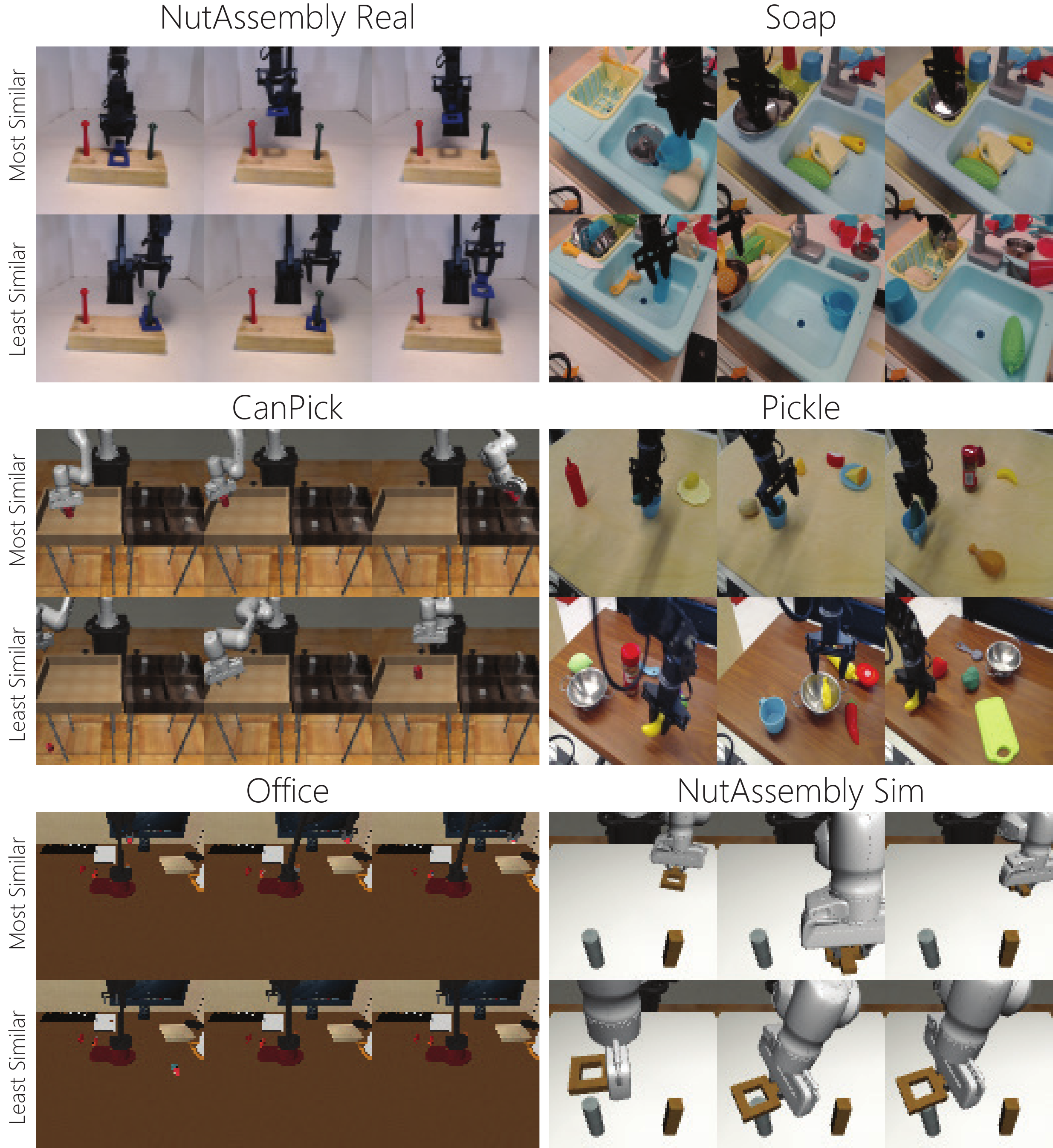}
    \caption{\small \textbf{Visualizing Retrieved Data for All Environments}: We show samples from the top 25 and the bottom 25 transitions when we apply our method to the offline data collected in each environment.}
    \vspace{-0.3cm}
    \label{app:allembeddings}
\end{figure}

Figure \ref{app:allembeddings} visualizes some of the selected and rejected data while running behavior retrieval on all of our six environments. As can be seen, the most similar transitions have correspondences to the desired task, and the least similar transitions typically contain irrelevant states. We note that for most environments, the grasping process is found in the top 25 transitions. This is desirable, as it shows that our method can make use of shared sub-trajectories. 
\section{Additional Experiments}

\subsection{Can \algocap~Perform Task Stitching?}
\label{exp:e5}

In our next experiment, we seek to understand if \algo~is capable of performing task-stitching in the simulated WidowX office environment. That is, we consider the case where $\dpr$ only contains short-horizon behaviors (e.g. picking a single object and placing it in a container), but test on a target task $\dt$ which requires picking and placing two objects in sequence. 
We see in Table~\ref{stitching} that in this setting \algo~is the only method able to achieve a success rate above 10\%. \algo~is able to pull the useful transitions from each of the short horizon trajectories in the offline data, and jointly finetune on it with $\dt$ to more effectively tackle the sequential task. 

\begin{table}[!h]
  \centering %
  \begin{tabular}{ccccc} 
    \toprule     
     $\dt$ & ($\dpr + \dt$) & GCBC $\dpr$ & GCBC $\dpr$  & Ours 
     \\
      & &  &  +FT $\dt$ & 
     \\
    \midrule
      $5\pm 3\%$ & $0$ & $0$ & $3\pm 2\%$ &\textbf{20$\pm$ 2\%} \\
    \bottomrule
  \end{tabular}
  \caption{\small \textbf{Robustness to Task Horizon.} We study if \algo~can perform task-stitching, that is, leveraging short horizon trajectories in $\dpr$ to learn a sequential task in $\dt$. We see that \algo~is the only method to get a success rate that is not near zero, due to it's ability to pull the relevant parts of short-horizon trajectories help learn the multi-step task.}
  \label{stitching}
  \vspace{-0.3cm}
\end{table}

\subsection{Can \algocap~Be Combined with Other Pre-Training Methods?}
\label{exp:e7}

In this experiment, we investigate if \algo~can be combined with other methods than vanilla behavior cloning for pre-training policies. Concretely, we combine behavior retrieval finetuning with the LMP approach proposed in \citet{Lynch2019LearningLP}, which learns a skill embedding space by encoding trajectories which the policy is then conditioned on. In Table~\ref{lmp} we look at success rates in the Robosuite CanPick environment, comparing LMP trained on $\dpr$, LMP trained on $\dpr$ and naively finetuned on $\dt$, and LMP trained on $\dpr$ and finetuned with \algo~on $\dpr$. We observe that naively finetuning the LMP policy on $\dt$ helps performance slightly, but using \algo~finetuning leads to more than a 10\% improvement in success rate. This suggests that \algo as a finetuning method can be naturally combined with different policy training approaches.

\begin{table}[!h]
  \centering %
  \begin{tabular}{ccc} 
    \toprule     
     LMP $(\dpr)$ & LMP $(\dpr)$ + FT $(\dt)$  & LMP $(\dpr)$ + FT (Ours)
     \\
    \midrule
      $55\pm 2\%$ & $56\pm 2\%$ &\textbf{67$\pm$ 4\%} \\
    \bottomrule
  \end{tabular}
  \caption{\small \textbf{Combining \algocap~with LMP.} We study if \algo~can be combined with other policy pre-trainined methods like LMP \cite{Lynch2019LearningLP}. We find that it can, producing over a 10\% improvement in success rate.}
  \label{lmp}
  \vspace{-0.3cm}
\end{table}

\subsection{Using the Full Bridge Dataset.}
\label{exp:e8}

In the original Pickle and Soap real robot tasks above, we set $\dpr$ to only be a subset of the full Bridge Dataset that was captured in visually similar scenes to the target environment. As a result, $\dpr$ only consisted of 10-20\% of the full Bridge Dataset. To test if \algo~can handle the full, more visually diverse dataset, we ran an additional experiment on the task of picking and placing a cup, where $\dpr$ consisted of the full Bridge Dataset, including mostly data from different scenes. We found that out of 20 trials, \algo~had a \textbf{55\%} success rate at the task, while $\dt$ only has a 30\% success rate, and $\dpr + \dt$ has a 35\% success rate. Thus, even when only a small portion of a visually diverse $\dpr$ is relevant to the task, with the correct choice of $\delta$ (0.9 in this case), \algo~can still provide significant benefits to performance.

\subsection{Failure Mode Analysis}
\label{exp:e6}

\begin{table}[!h]
  \centering %
  \begin{tabular}{c|ccc} 
    \toprule     
    & Grasp & Intermediate & Alignment / Place \\
    \midrule
    Ground Truth & 52\% & 0\% & 58\%  \\
    Ours & 54\% & 2\% & 44\%  \\
    Task-Specific Only & 60\% & 9\%  & 31\% \\
    All Data & 22\% & 31\% & 47\%  \\
    GC & 83\% & 5\%  & 12\% \\
    GC + Finetuned & 15\% & 0\%  & 85\% \\
    \bottomrule
  \end{tabular}
  \caption{\small \textbf{Failure Mode Analysis}. Averaged across all of our real robot tasks (40 trials each), we measure the percentage of each type of failure for each of the methods including our own. We observe that the distribution of failures is similar to Ground Truth, suggesting that with effective retrieval, methods tend to only fail at the challenging bottleneck regions of the task (grasping and placing).}
  \label{failuremode}
  \vspace{-0.3cm}
\end{table}

For the real robot experiments, we kept track of the types and frequencies of failures on the tasks. We grouped these failures into three categories relevant to the three tasks we consider: grasp, intermediate, and alignment / place. Intermediate failures include anything between the grasping of the object and the final alignment process, like the premature release of the object.

We plot the percentage of each failure type for each method in Table~\ref{failuremode}, averaged across the three tasks and 40 trials.

Averaged across the three tasks, the failure distribution of \algo~is comparable to the ground truth. Namely, most of the failures are at the bottleneck regions of grasping and placing. When all the data is included, the robot experiences more intermediate failures, as the offline dataset contains irrelevant and even adversarial demonstrations, like in the case of the two pegs in Nut Assembly.

Finally, with the goal-conditioned models, the significant distribution shift between the offline and task-specific demonstrations often yielded policies that failed to successfully grasp the object. The grasping is fixed with fine-tuning, although perhaps due to the complexity of the goal state, the robot struggles with the higher precision alignment part of the task.

\end{document}